\documentclass[10pt, journal,web]{article}
\usepackage{times}
\usepackage{subfig}
\usepackage[utf8]{inputenc}
\usepackage{utfsym}
\usepackage{multirow}
\usepackage[normalem]{ulem}
\usepackage{amsmath,amssymb,amsfonts}
\usepackage{algorithmic}
\useunder{\uline}{\ul}{}
\usepackage{url}
\def\BibTeX{{\rm B\kern-.05em{\sc i\kern-.025em b}\kern-.08em
    T\kern-.1667em\lower.7ex\hbox{E}\kern-.125emX}}
\usepackage{graphicx}
\usepackage{authblk}

\usepackage{longtable}
\usepackage{pdflscape}
\usepackage[T1]{fontenc}

\usepackage{xcolor}
\usepackage{hyperref}
\usepackage{booktabs}
\hypersetup{
   colorlinks=true,
    linkcolor=blue,
   citecolor=blue,
    urlcolor=blue
}
\usepackage{makecell}
\usepackage{tabularx}
\usepackage{enumitem}

\makeatletter
\renewcommand\AB@affilsepx{, \protect\Affilfont}
\makeatother
\usepackage{tabularx}
\providecommand{\keywords}[1]
{
  \small	
  \textbf{\textit{Keywords---}} #1
}

\begin{document}

\title{\textbf{Agentic AI: A Comprehensive Survey of Architectures, Applications, and Future Directions}}
\author[1, 3, 4]{Mohamad Abou Ali}

\author[1, 2]{Fadi Dornaika\thanks{Corresponding author}}
\affil[1]{\textit{University of the Basque Country}}
\affil[2]{\textit{IKERBASQUE}}
\affil[3]{\textit{Lebanese International University (LIU)}}
\affil[4]{\textit{The International University of Beirut}}

\affil[ ]{

\small\texttt{mohamad.abouali01@liu.edu.lb, fadi.dornaika@ehu.eus}}
\date{}
\maketitle
\begin{abstract}
Agentic AI represents a transformative shift in artificial intelligence, but its rapid advancement has led to a fragmented understanding, often conflating modern neural systems with outdated symbolic models—a practice known as \emph{conceptual retrofitting}. This survey cuts through this confusion by introducing a novel \textbf{dual-paradigm framework} that categorizes agentic systems into two distinct lineages: the \textbf{Symbolic/Classical} (relying on algorithmic planning and persistent state) and the \textbf{Neural/Generative} (leveraging stochastic generation and prompt-driven orchestration). Through a systematic PRISMA-based review of 90 studies (2018–2025), we provide a comprehensive analysis structured around this framework across three dimensions: (1) the theoretical foundations and architectural principles defining each paradigm; (2) domain-specific implementations in healthcare, finance, and robotics, demonstrating how application constraints dictate paradigm selection; and (3) paradigm-specific ethical and governance challenges, revealing divergent risks and mitigation strategies. Our analysis reveals that the choice of paradigm is strategic: symbolic systems dominate safety-critical domains (e.g., healthcare), while neural systems prevail in adaptive, data-rich environments (e.g., finance). Furthermore, we identify critical research gaps, including a significant deficit in governance models for symbolic systems and a pressing need for hybrid neuro-symbolic architectures. The findings culminate in a strategic roadmap arguing that the future of Agentic AI lies not in the dominance of one paradigm, but in their intentional integration to create systems that are both \emph{adaptable} and \emph{reliable}. This work provides the essential conceptual toolkit to guide future research, development, and policy toward robust and trustworthy hybrid intelligent systems.
\end{abstract}

 \hspace{10pt}

\section{Introduction}

The field of Artificial Intelligence (AI) is undergoing a paradigm shift from the development of passive, task-specific tools toward the engineering of autonomous systems that exhibit genuine agency. Modern agentic AI systems \cite{wissuchek2025exploring, viswanathan2025agenticframework} are defined by capabilities such as proactive planning, contextual memory, sophisticated tool use, and the ability to adapt their behavior based on environmental feedback. These systems operate not as mere solvers but as collaborative partners, capable of dynamically perceiving complex environments, reasoning about abstract goals, and orchestrating sequences of actions—either independently or as part of a sophisticated multi-agent ecosystem~\cite{xie2024large, du2025survey}.

To establish a precise conceptual foundation, we distinguish between the field's core concepts. An \textit{AI Agent} (or a \textit{Single-Agent System}) is a self-contained autonomous system designed to accomplish a goal. It operates primarily in isolation, though it may interact with tools and APIs. Its agency is defined by its \textit{autonomy}, \textit{proactivity}, and its ability to complete a task from start to finish independently.

For example, a single, powerful \textit{LLM-based (Large Language Model-based)} agent tasked with ``Write a full project proposal for a new mobile app'' would autonomously break down the task, conduct research, write the sections, and format the final document.

In contrast, \textit{Agentic AI} is the broader field and architectural approach concerned with creating systems that exhibit agency. Crucially, this often involves the orchestration of \textit{Multi-Agent Systems (MAS)}, where multiple specialized agents work together, coordinating and communicating to solve problems that are too complex for a single agent.

For example, an Agentic AI system designed for the same task would employ a team of specialized agents: a \textit{Project Manager Agent} to break the goal into tasks, a \textit{Researcher Agent} to gather market data, a \textit{Writer Agent} to draft content, and a \textit{Quality Assurance Agent} to review the output. Their collaborative workflow is the embodiment of Agentic AI.

In summary, one can conceptualize an \textit{AI Agent} as a single, sophisticated worker, while \textit{Agentic AI} represents the principle of leveraging agency, frequently by architecting and managing an entire team of such workers.

This rapid evolution, however, has led to a fragmented and often anachronistic understanding of the field. A critical issue identified in prior reviews is \textit{conceptual retrofitting}—the misapplication of classical symbolic frameworks (e.g., Belief–Desire–Intention (BDI) \cite{archibald2024quantitative}, \textit{perceive–plan–act–reflect (PPAR)} loops \cite{zeng2024perceive, erdogan2025plan}) to describe modern systems built on \textit{large language models (LLMs)} \cite{plaat2025agentic}, which operate on fundamentally different principles of stochastic generation and prompt-driven orchestration. This practice obscures the true operational mechanics of LLM-based agents \cite{gabison2025inherent, Wang2024Survey, Zhao2023Indepth, Chen2024Survey} and creates a false sense of continuity between incompatible architectural paradigms, whether applied to a single complex agent or a coordinated MAS.

\begin{landscape} \begin{figure}[p] \centering \includegraphics[width=\linewidth, height=1.5\textheight, keepaspectratio]{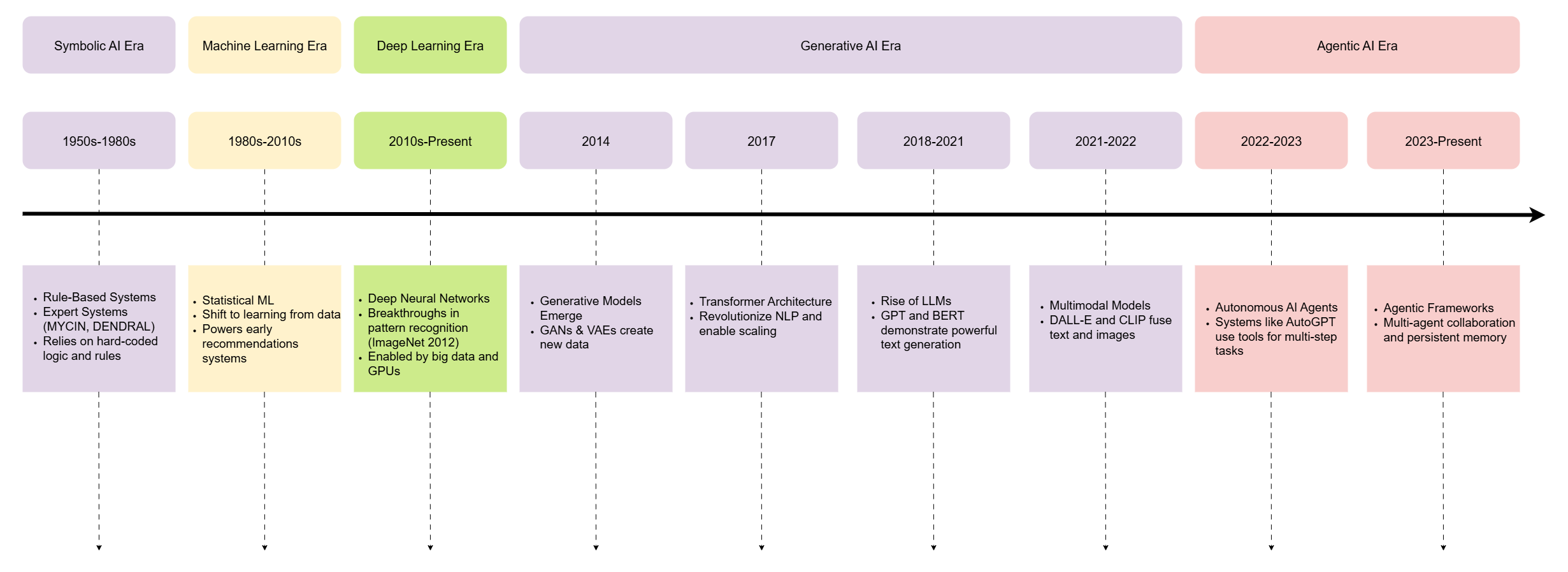} \caption{Historical Evolution of AI Paradigms: This timeline charts the key breakthroughs and eras in AI, from early symbolic systems to the modern agentic era. It highlights the Transformer architecture as the pivotal enabling technology for large language models (LLMs), which in turn powered the generative AI revolution and provided the substrate for contemporary agentic systems.} \label{fig:historical_timeline} \end{figure} \end{landscape}

This paper addresses these gaps by first establishing a clear historical context (Figure~\ref{fig:historical_timeline}), which delineates the evolution of AI through five distinct but overlapping eras.  

The \textbf{Symbolic AI Era (1950s--1980s)} \cite{Liang2025} established the foundational ambition of artificial intelligence, grounded in logic and explicit human knowledge. This period was dominated by rule-based systems and expert systems such as MYCIN and DENDRAL \cite{Swartout1985}, which operated on carefully hand-crafted symbolic rules. Intelligence was conceived as a top-down, deductive process, representing the purest form of the symbolic paradigm.  

The \textbf{Machine Learning (ML) Era (1980s--2010s)} \cite{Thomas2020,Nithya2023,Trigka2025} marked a pivotal shift away from hard-coded logic toward systems that could learn from data. While still heavily dependent on human-engineered features, this period introduced statistical ML models such as Support Vector Machines and decision trees, which powered applications ranging from classification to recommendation. It was a transitional stage that moved the field away from pure symbolism but still lacked the automated feature learning that would define subsequent eras.  

The arrival of the \textbf{Deep Learning Era (2010s--Present)} \cite{Hatcher2018,Alom2019,Dong2021,Talaei2023,Chhabra2023} was catalyzed by the confluence of increased compute power and large datasets. Deep neural networks, including convolutional and recurrent architectures, enabled systems to automatically learn hierarchical representations from raw data. This era revolutionized pattern recognition in vision, speech, and text, breaking longstanding barriers in perception. Yet, despite their power, these models largely functioned as sophisticated pattern classifiers rather than autonomous agents.  

Out of this foundation emerged the \textbf{Generative AI Era (2014--Present)} \cite{Sakirin2023,Anandhi2025,Sengar2024,Surbakti2025,Zhang2025}, fueled by advances in generative modeling. Early breakthroughs such as Generative Adversarial Networks were soon eclipsed by the introduction of the Transformer architecture in 2017, which enabled the scaling of large language models (LLMs) such as GPT and BERT. These systems moved beyond perception to generation, producing coherent text, code, and media. In doing so, they provided the essential substrate---a powerful, general-purpose statistical reasoner---that made modern agentic AI feasible.  

Finally, the \textbf{Agentic AI Era (2022--Present)} represents the current frontier, where the generative capabilities of LLMs are harnessed for action and autonomy. This era is characterized by the rise of AI agents \cite{Durante2024,Masterman2024,Piccialli2025} such as AutoGPT, which can pursue goals through planning and tool use. Increasingly, these agents evolve into multi-agent systems \cite{Acharya2025,Viswanathan2025,Plaat2025,Schneider2025,Hosseini2025}, exemplified by frameworks like CrewAI and AutoGen, where specialized roles and orchestrated collaboration enable teams of agents to tackle complex problems. In contrast to the algorithmic deliberation of the symbolic paradigm, this stage is defined by the neural paradigm, where agency emerges from the stochastic orchestration of generative models.  

\begin{figure}[htbp]
    \centering
    \includegraphics[width=0.95\textwidth]{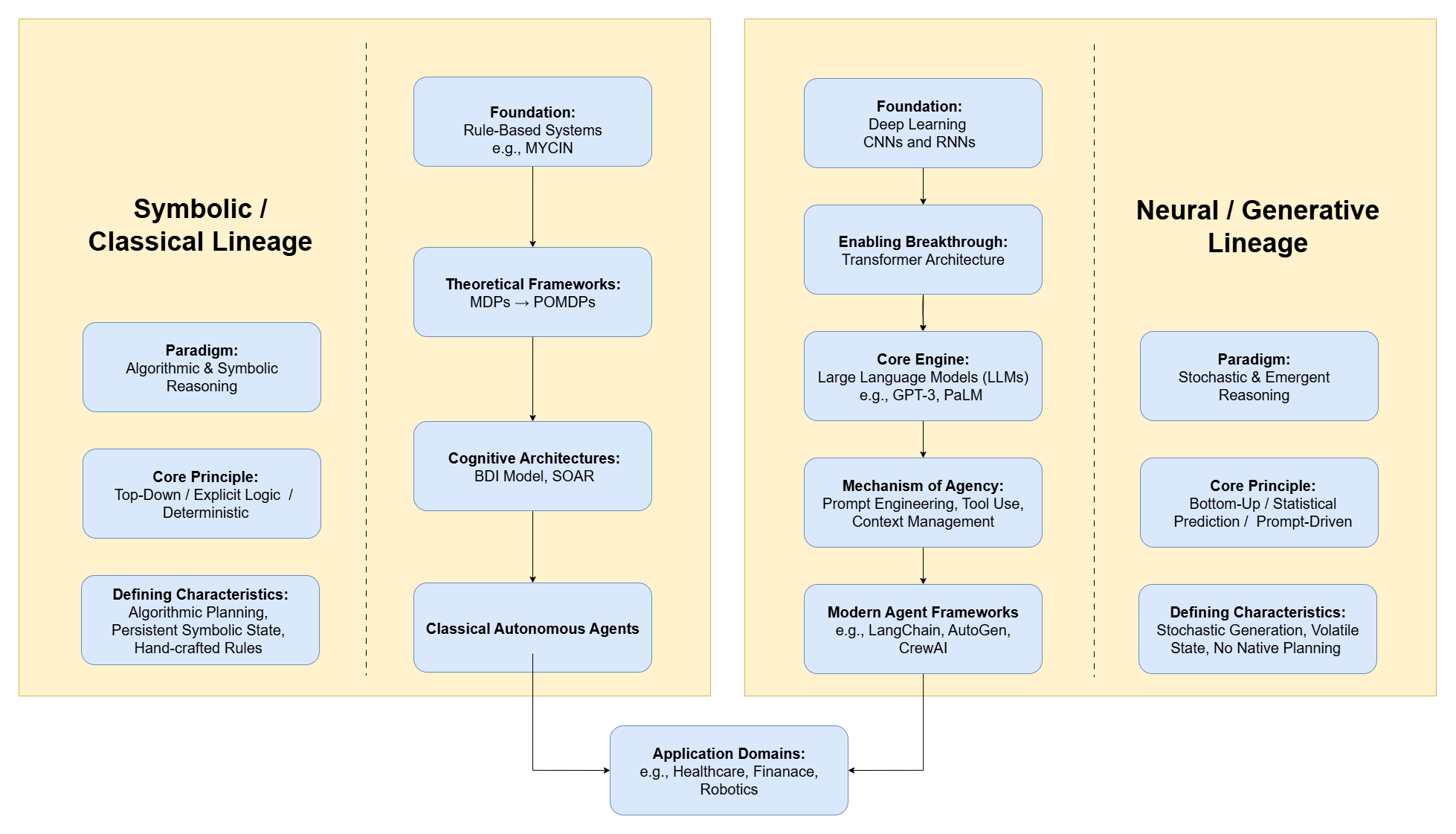} 
    \caption{Conceptual Framework of Agentic AI's Dual Lineages. This taxonomy resolves conceptual retrofitting by distinguishing the Symbolic/Classical lineage (left), defined by algorithmic planning and persistent state, from the Neural/Generative lineage (right), defined by stochastic generation and prompt-driven orchestration. While both paradigms target similar applications, their underlying mechanisms are fundamentally incompatible. This framework provides the analytical structure for this survey.}
    \label{fig:conceptual_framework}
\end{figure}

This chronological progression provides essential context but also reveals a critical conceptual schism. The agentic AI era is not simply a linear descendant of symbolic AI but is instead built upon a completely different architectural foundation. To address this, we introduce a novel conceptual framework (Figure~\ref{fig:conceptual_framework}) designed to prevent retrospective conflation by clearly distinguishing the symbolic and neural lineages of agentic AI. This dual-axis taxonomy provides the unified lens necessary to rigorously analyze the field's theoretical underpinnings, architectural innovations, and practical deployments.  

The journey to modern agentic AI is best understood through its historical progression, as detailed in Figure~\ref{fig:historical_timeline}. This evolution moved from the deterministic, rule-based systems of the symbolic era through the data-driven revolutions of machine learning and deep learning, culminating in the transformative advent of large language models (LLMs) \cite{zhao2023survey,wan2024efficient} and generative AI.

However, a chronological account is insufficient for analytical rigor. The central challenge in current discourse is the conceptual retrofitting of modern, neural agentic architectures into the frameworks of the symbolic era. To resolve this, we propose a dual-paradigm taxonomy in Figure~\ref{fig:conceptual_framework}. This framework categorizes agentic systems along two independent dimensions: their \textbf{Architectural Paradigm} (Symbolic vs. Neural) and their \textbf{Degree of Agency \& Coordination} (Single-Agent vs. Multi-Agent). This model is designed not to show evolution, but to provide a clear analytical structure for classification and comparison, ensuring systems are evaluated on their own operational terms.

This review is structured around this framework to synthesize three critically interconnected layers:

The first layer encompasses the \textbf{Theoretical Foundations}, including core principles of autonomy and agency~\cite{kolt2025governing}, and decision-making models like Markov Decision Processes (MDPs) and Partially Observable MDPs (POMDPs)~\cite{kaelbling1998planning, Chades2021-primer}. It is crucial to note that these models provide a theoretical language for describing agency that originated in the \textit{Symbolic paradigm}, but modern systems \textit{implement} these concepts in entirely new ways.

The second layer analyzes \textbf{Architectural Frameworks}, focusing on the modern infrastructures powering the \textit{Neural paradigm}. We examine systems like LangChain~\cite{mavroudis2024langchain}, AutoGen, and CrewAI, which achieve agency through mechanisms like prompt chaining, conversation orchestration, and dynamic context management—a clear departure from the symbolic planning of the classical lineage.

The third layer investigates \textbf{Application Domains}, exploring the practical deployment of agentic systems across fields such as healthcare~\cite{singh2024revolutionizing}, finance~\cite{chandrashekar2025survey}, scientific discovery~\cite{gridach2025agentic}, and legal reasoning~\cite{magesh2024hallucination}. Our framework allows us to map these applications to the appropriate paradigm and analyze their unique implementation challenges.

\vspace{-2mm}
\subsection{\textbf{Current Surveys Gaps and Contributions}}

The current discourse on agentic AI suffers from the conceptual retrofitting illustrated in Figure~\ref{fig:conceptual_framework}. Classical AI frameworks, such as the BDI model or perceive–plan–act–reflect (PPAR) loops, are often rhetorically applied but are fundamentally mismatched to the stochastic, non-symbolic, and context-driven nature of LLM-based agents~\cite{archibald2024quantitative}. Furthermore, existing reviews are often narrow in scope, lacking empirical comparisons or integrated governance insights. As summarized in Table~\ref{tab:prior_surveys}, current literature leaves substantial gaps in understanding the field's current state.

\begin{table}[h]
\centering
\caption{Summary of Prior Surveys on Agentic AI}
\begin{tabularx}{\textwidth}{
    >{\hsize=0.7\hsize}X
    >{\hsize=0.8\hsize}X
    >{\hsize=1\hsize}X
    >{\hsize=1\hsize}X
}
\toprule
\textbf{Reference} & \textbf{Focus} & \textbf{Key Contributions} & \textbf{Limitations} \\
\midrule
Plaat et al. (2025) \cite{plaat2025agentic} & Agentic LLMs & Reasoning-Acting-Interacting taxonomy & Limited empirical validation; no evolutionary context \\
Schneider (2025) \cite{schneider2025generative} & GenAI to Agentic shift & Conceptual framework for autonomy & No performance metrics; ignores architectural mechanisms \\
Acharya et al. (2025) \cite{acharya2025agentic} & Foundational methods & Combined RL with cognitive architectures & Scalability not addressed; overlooks LLM-based paradigms \\
Gridach et al. (2025) \cite{gridach2025agentic} & Scientific discovery & Tools for autonomous research workflows & No governance discussion; isolated application view \\
Hosseini \& Seilani (2025) \cite{hosseini2025role} & Enterprise strategy & Agentic design for organizational alignment & Lack of technical depth; no architectural analysis \\
Ozman (2025) \cite{sapkota2025ai} & Business operations & Systematic review methodology & Missing benchmark comparisons; no unifying framework \\
\bottomrule
\end{tabularx}
\label{tab:prior_surveys}
\end{table}

This review directly addresses these limitations through four integrated contributions:

\begin{enumerate}[label=\textbf{\arabic*.}, leftmargin=*, itemsep=1em]
    \item \textbf{A Novel Dual-Paradigm Taxonomy:} We introduce and employ the framework in Figure~\ref{fig:conceptual_framework} as our primary analytical tool, explicitly distinguishing symbolic and neural lineages to prevent conceptual retrofitting and enable accurate system classification.

    \item \textbf{Architectural Clarification:} We demystify the operational principles of modern neural frameworks (Section~\ref{sec:literature}), explaining how they achieve agency through mechanisms like prompt chaining and conversation orchestration, rather than symbolic planning.

    \item \textbf{Empirical Mapping:} We conduct a systematic PRISMA-based literature review of 90 studies, categorizing them using our dual-paradigm framework to trace research trends and evaluate architectures by their appropriate standards.

    \item \textbf{Governance Anchoring:} We embed ethical, accountability, and alignment challenges within each paradigm of our taxonomy to ensure that safety considerations are discussed in the correct technological context (Section~\ref{sec:ethics}).
\end{enumerate}

\vspace{-2mm}
\subsection{\textbf{Structure of the Paper}}

To guide the reader through our analysis, the paper is structured to logically develop the argument for a dual-paradigm understanding of Agentic AI. We begin by establishing the necessary theoretical context in Section~\ref{sec:foundations}, which explores the foundations of agency and introduces our core taxonomic framework. Section~\ref{sec:methodology} then details the systematic methodology underpinning our literature review.

The subsequent sections apply this framework to analyze the field: Section~\ref{sec:literature} reviews key architectural frameworks through our taxonomic lens, and Section~\ref{sec:applications} examines how different application domains influence paradigm selection. Section~\ref{sec:taxonomy} presents a comprehensive paradigm-aware taxonomy of the literature, serving as a foundational reference and key output of our review. Section~\ref{sec:ethics} investigates the paradigm-specific nature of ethical and governance challenges, leading directly into Section~\ref{sec:gaps}, which outlines the critical research gaps identified by our analysis.

The final sections synthesize our findings and look forward. Section~\ref{sec:future} then charts an actionable research roadmap toward hybrid intelligence, building directly upon both the identified gaps and our stated contributions.  Finally, Section~\ref{sec:conclusion} provides a final synthesis of our findings and their implications for the field.

This structure is designed to first equip the reader with the necessary conceptual tools, then systematically analyze the landscape, and conclude by synthesizing the insights into a coherent vision for the future of Agentic AI.

\section{Theoretical Foundations: Mapping the Dual Lineages of Agentic Intelligence}
\label{sec:foundations}

The architectural history of agentic AI is not a linear progression but a branching into two distinct paradigms, as defined by our conceptual framework (Figure~\ref{fig:conceptual_framework}). This section delineates the theoretical and cognitive groundwork for both the \textbf{Symbolic/Classical} and \textbf{Neural/Generative} lineages, clarifying their foundational principles and highlighting the paradigm shift that separates them.

\subsection{Core Principles of Autonomy and Agency}
The conceptual language for describing agency originated within the symbolic paradigm. The foundational constructs of \textit{autonomy} and \textit{agency} are essential for both lineages, though they are implemented in fundamentally different ways. Autonomy refers to a system’s ability to operate independently, free from direct human intervention, whereas agency encapsulates the notion of goal-directed behavior that incorporates intention, contextual awareness, and decision-making capabilities \cite{patel2020demonstration, kolt2025governing}. Agentic AI synthesizes these traits by initiating tasks, dynamically ranking goals, monitoring progress, and adjusting behavior through feedback loops \cite{trencsenyi2025influence}.

These mechanisms parallel human executive functions such as planning, inhibition, and cognitive flexibility. They provide the high-level descriptive framework for intelligent behavior, which both symbolic and neural systems aim to achieve through divergent mechanisms.

\subsection{The Symbolic Lineage: Algorithmic Decision-Making}
The symbolic lineage is characterized by explicit logic, algorithmic planning, and deterministic or probabilistic models. Its evolution provides the theoretical bedrock for pre-LLM autonomous systems.

\subsubsection{Markov Decision Processes (MDPs)}
MDPs provide the mathematical scaffolding for modeling environments with full state information \cite{lu2019general, lu2023decision}, a hallmark of early symbolic and classical statistical AI. An MDP is defined by a tuple (S, A, P, R), representing states, actions, transition probabilities, and rewards. These systems operate effectively in deterministic, rule-based domains but lack the capacity for robust reasoning under uncertainty, anchoring them firmly in the symbolic paradigm.

\subsubsection{Partially Observable MDPs (POMDPs)}
\begin{figure}[htbp]
\centering
\includegraphics[width=0.9\textwidth]{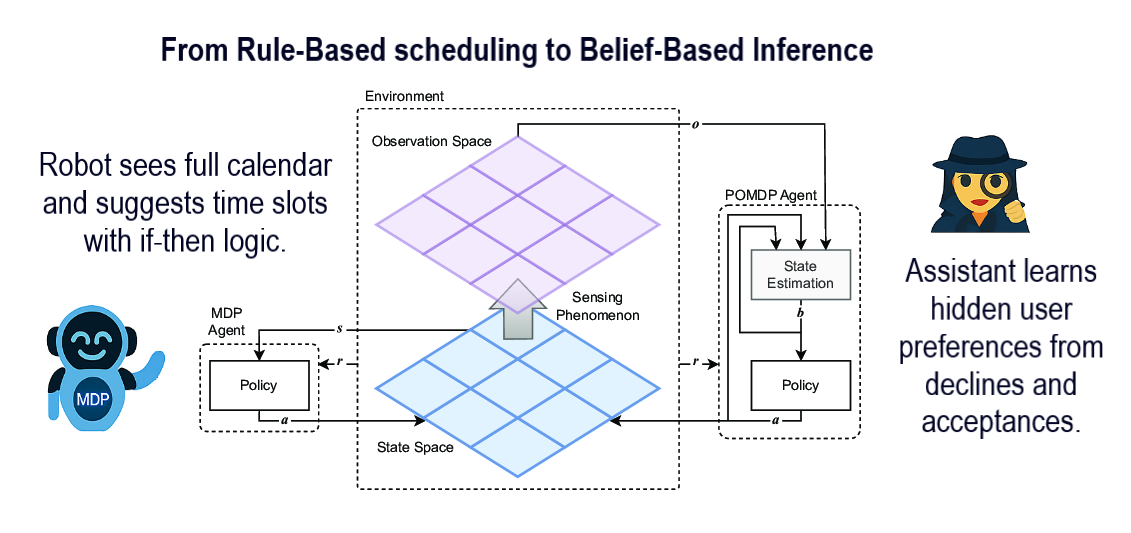}
\caption{Classical symbolic reasoning: Comparison between a rule-based MDP scheduler (left) and a belief-based POMDP assistant (right). The MDP agent relies on explicit calendar states and deterministic policies, while the POMDP agent infers hidden user preferences from behavioral feedback. Both represent the symbolic paradigm's approach to decision-making.}
\label{fig:mdp_pomdp_comparison}
\end{figure}

POMDPs extend MDPs by introducing probabilistic \textit{belief states} to handle environments where the agent has incomplete information \cite{rozek2024partially, lu2024rethinking}. This was a key advancement, allowing symbolic agents to infer hidden states through observation and enabling more adaptive behavior. However, as illustrated in Figure~\ref{fig:mdp_pomdp_comparison}, this is still a form of algorithmic state estimation. The significant computational overhead of belief tracking limits their scalability and real-world application \cite{frering2025integrating, gillen2020explicitly}, a fundamental constraint of the symbolic approach.

\subsubsection{Cognitive Architectures: BDI and SOAR}
Cognitive architectures like Belief-Desire-Intention (BDI) and SOAR represent the pinnacle of the symbolic paradigm's attempt to engineer agency. They explicitly model internal states and processes, as summarized in Table~\ref{tab:cognitive_mapping}. These systems directly implement a perceive-plan-act-reflect loop using symbolic representations, making them powerful but brittle and difficult to scale to complex, real-world environments. Their relationship to human cognitive functions is a direct, top-down mapping of symbolic logic.

\begin{table}[htbp]
\centering
\caption{Mapping Human Cognitive Functions to Symbolic AI Modules}
\label{tab:cognitive_mapping}
\begin{tabularx}{\textwidth}{lXX}
\toprule
\textbf{Component} & \textbf{Human Function} & \textbf{Symbolic AI Parallel} \\
\midrule
\textbf{Belief Module} & Working Memory & Symbolic Knowledge Base / World Model \\
\textbf{Desire Module} & Motivation & Goal Stack / Utility Function \\
\textbf{Intention Module} & Executive Control & Action Policy / Planner \\
\textbf{Meta-cognition Layer} & Self-reflection, Error Monitoring & Monitor / Replan Loop \\
\bottomrule
\end{tabularx}
\end{table}

\subsection{The Neural Lineage: Statistical Learning and Emergent Reasoning}
The neural lineage is built on a foundation of statistical learning from data, culminating in the generative capabilities of large language models (LLMs). Its progression is marked by a move away from explicit logic toward emergent, stochastic behavior.

\subsubsection{Deep Reinforcement Learning (DRL)}
\begin{figure}[htbp]
\centering
\includegraphics[width=0.85\textwidth]{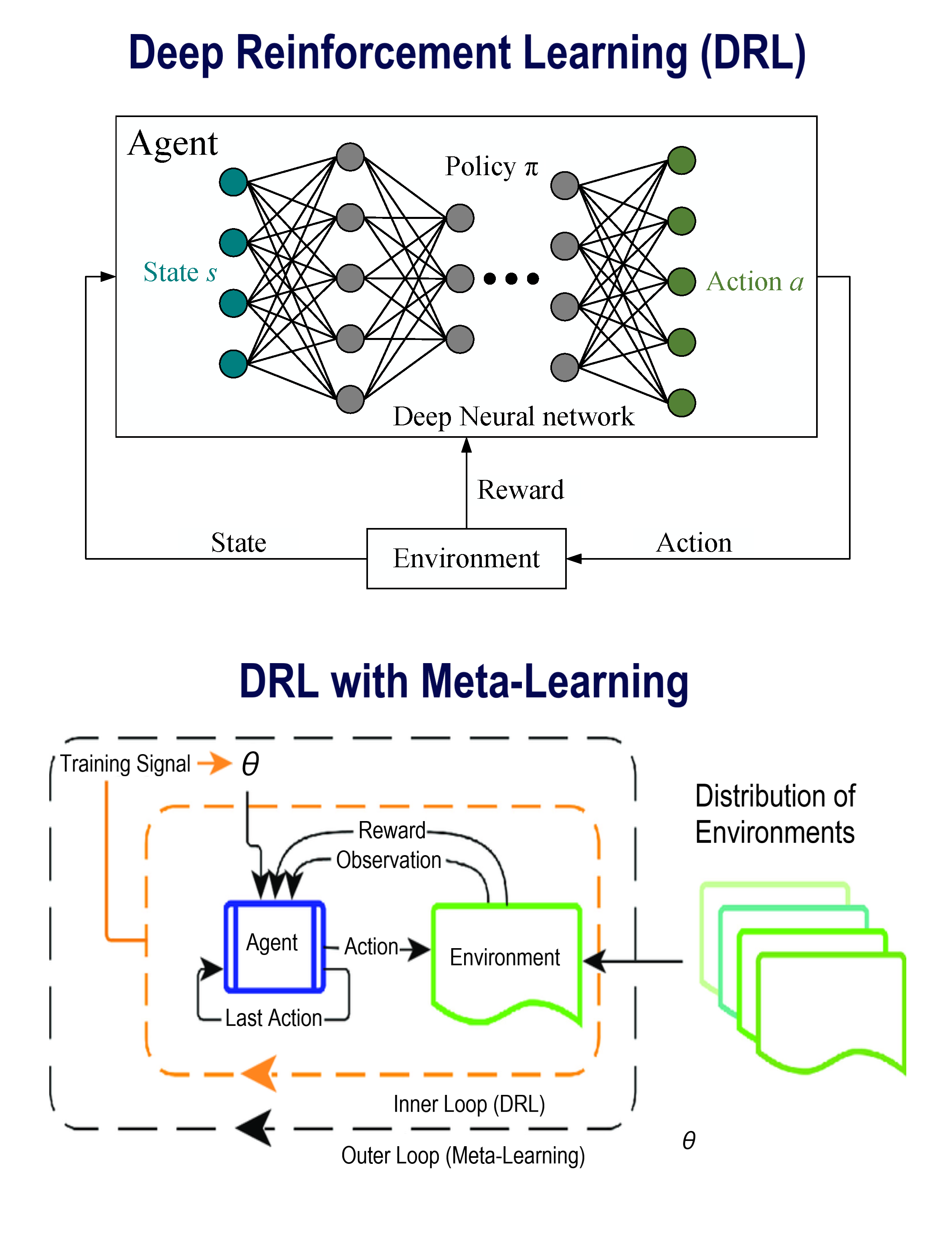}
\caption{The shift toward learned behavior: Architectural contrast between vanilla DRL (single-task optimization) and meta-DRL (dual-loop generalization). The latter improves adaptability across tasks through meta-optimization loops, moving from explicit programming toward learned, emergent capabilities.}
\label{fig:drl_comparison}
\end{figure}

Deep Reinforcement Learning (DRL) represents a critical transition. It scales learning to high-dimensional inputs (like images and text) using neural networks \cite{singh2025agentic, bodepudi2020agentic}. DRL agents learn policies directly from data, moving away from hand-crafted symbolic rules. Methods such as PPO allow for fine-grained behavioral optimization \cite{kumar2025proximal, yazid2023autonomous}. As shown in Figure~\ref{fig:drl_comparison}, advancements like meta-DRL introduced generalization across tasks, a precursor to the adaptability required for modern agency. DRL is a bridge, using neural networks to learn the policies that symbolic systems would have to be explicitly programmed with.

\subsubsection{The LLM Substrate and The Paradigm Shift}
The emergence of Large Language Models (LLMs) was not an evolution but a revolution that created the new neural paradigm. LLMs provided a powerful, general-purpose substrate for reasoning based on statistical prediction in a high-dimensional space of concepts. This enabled a fundamental architectural shift from designing cognitive agents to orchestrating generative pipelines.

Frameworks like LangChain, AutoGen, and CrewAI do not implement symbolic PPAR loops or BDI architectures. They represent a new paradigm of \textbf{LLM Orchestration}, where pre-trained models act as central executives that coordinate tasks through fundamentally different mechanisms, as detailed in Table~\ref{tab:orch}.

\begin{figure}[htbp]
\centering
\includegraphics[width=0.7\textwidth]{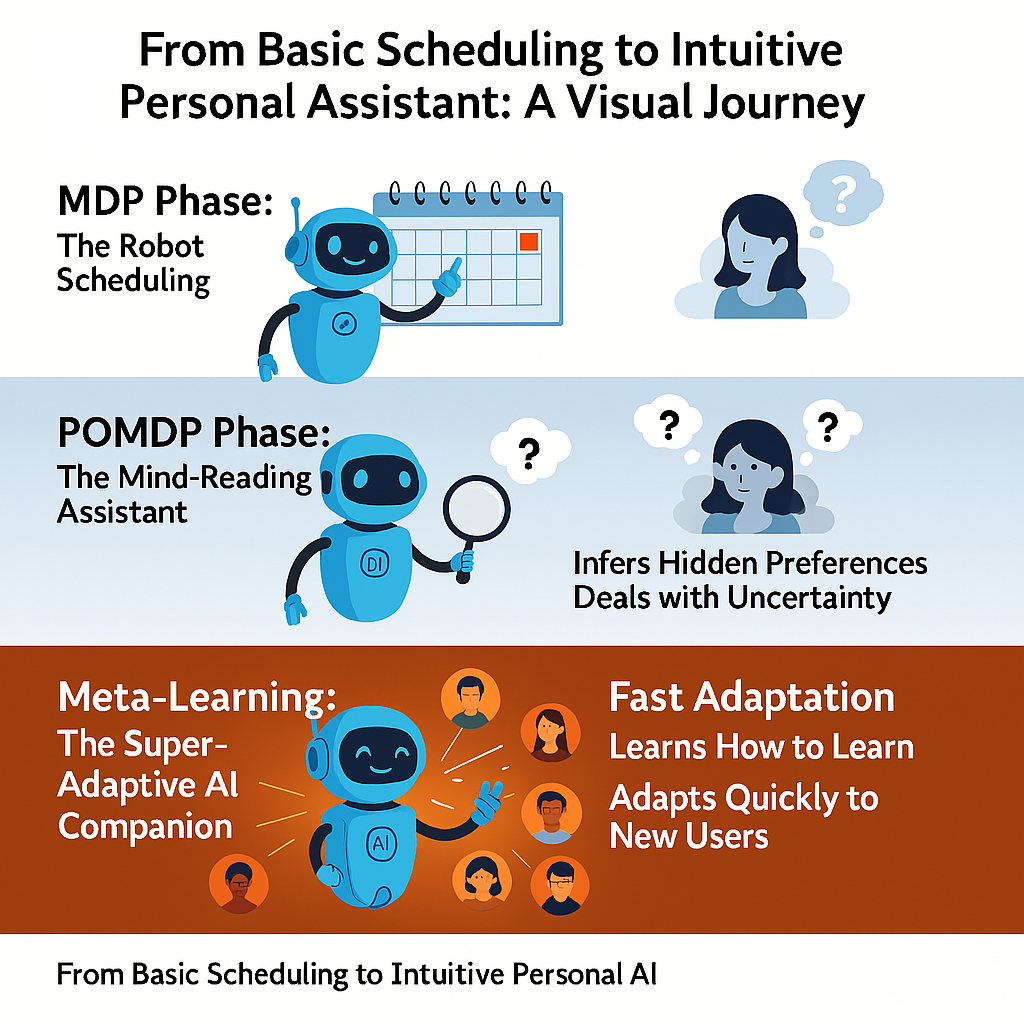}
\caption{The journey from symbolic to neural agency: The evolution of a personal assistant from a deterministic rule-based (MDP) system, to an uncertainty-aware (POMDP) system, and finally to a modern LLM-orchestrated agent. This journey bridges the two paradigms, ending with a system that exhibits intelligent behavior through entirely different mechanisms.}
\label{fig:assistant_evolution}
\end{figure}

This shift marks the definitive break from the symbolic tradition. Agency in the neural paradigm is an emergent property of prompt-driven orchestration, not a product of internal symbolic logic. The evolution of a personal assistant, depicted in Figure~\ref{fig:assistant_evolution}, culminates in this new architecture.

\begin{table}[h!]
\centering
\caption{Orchestration Mechanisms of Modern Neural Agentic Frameworks}
\label{tab:orch}
\begin{tabularx}{\textwidth}{p{2.5cm} p{2.5cm} p{5cm}}
\toprule
\textbf{Framework} & \textbf{Primary Mechanism} & \textbf{Functional Paradigm and Representative Applications} \\
\midrule
\textbf{LangChain}  \cite{mavroudis2024langchain, johnson2025langchain, gupta2024langchain, topsakal2023creating, taulli2025building} & Prompt Chaining & Orchestrates linear sequences of LLM calls and API tools. Replaces symbolic planning with stochastic generation of next steps. Applications: Multi-step workflow automations, automated medical reporting \cite{huh2023breast}. \\
\midrule
\textbf{AutoGen} \cite{wu2023autogen, dibia2025multi} & Multi-Agent Conversation & Facilitates structured dialogues between collaborative LLM agents. Replaces monolithic control with emergent problem-solving through conversation. Applications: Collaborative task solving, economic research coordination \cite{dawid2025agentic}. \\
\midrule
\textbf{CrewAI} \cite{venkadesh2024unlocking, duan2024exploration} & Role-Based Workflow & Assigns roles and goals to a team of agents, managing their interaction workflow. Replaces centralized scheduling with dynamic, role-driven process management. Applications: Market analysis and risk modeling \cite{chandrashekar2025survey}. \\
\midrule
\textbf{Semantic Kernel}  \cite{kothapalli2024integrating, meyer2024building, costea2025microsoft} & Plugin/Function Composition & Connects LLMs to pre-written code functions ("skills"). Replaces integrated actuation with stochastic planning of plugin sequences. Applications: Breaking down high-level user intents into executable skills. \\
\midrule
\textbf{LlamaIndex} \cite{gheorghiu2024building, ramirez2025accelerating, mozolevskyi2024comparative, braunschweiler2025evaluating} & Retrieval-Augmented Generation (RAG) & Provides sophisticated data connectors and indexing. Replaces internal symbolic knowledge bases with on-demand, external context retrieval. Applications: Financial sentiment analysis \cite{konstantinidis2024finllama}, enhancing information retrieval for research \cite{kommineni2025harnessing}. \\
\bottomrule
\end{tabularx}
\end{table}

\subsection{Multi-Agent Orchestration: The Pinnacle of the Neural Paradigm}
\begin{figure}[htbp]
\centering
\includegraphics[width=\textwidth]{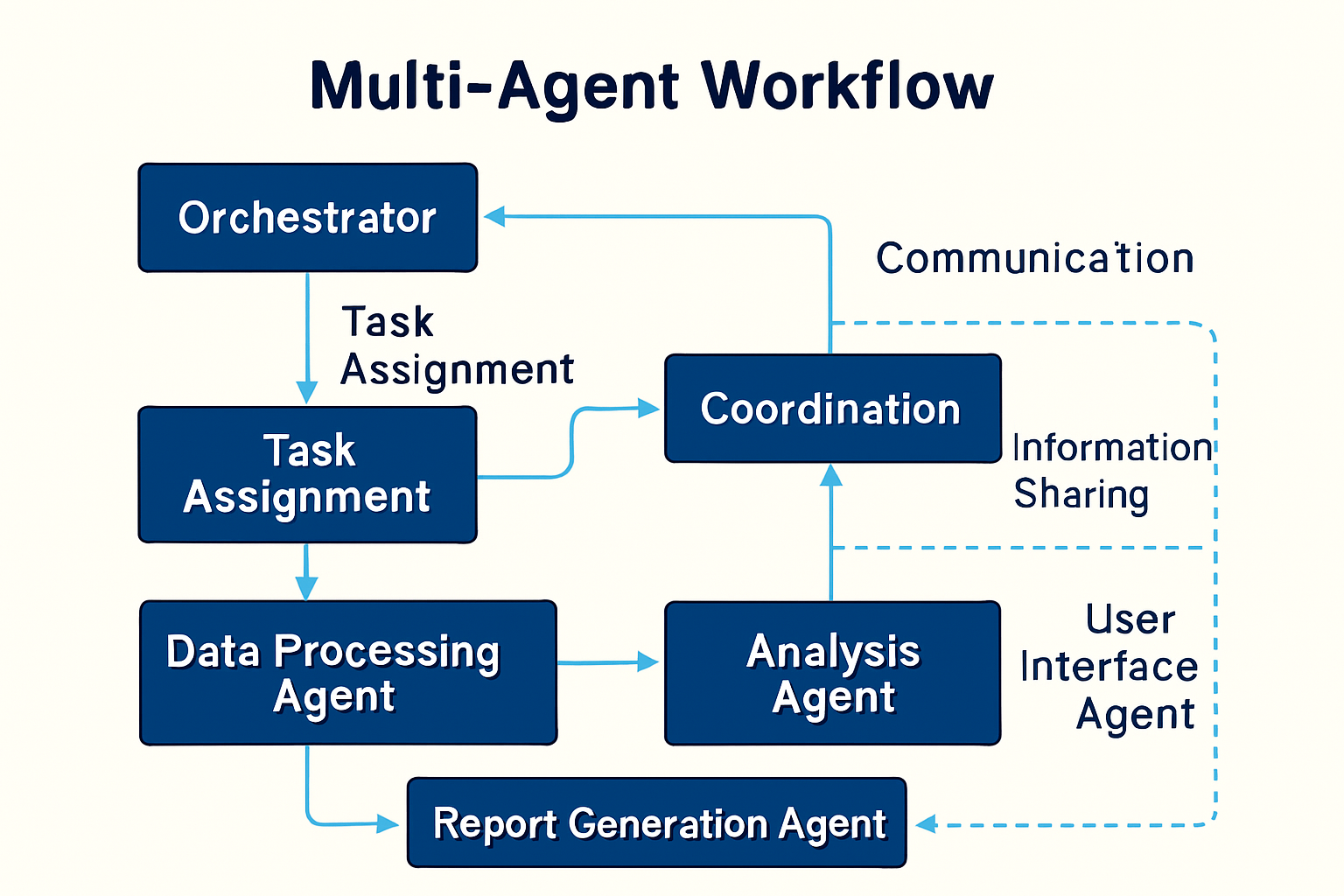}
\caption{The architecture of the neural paradigm: Multi-Agent Orchestration in modern AI systems. This schematic illustrates the operational paradigm of neural systems. A central orchestrator (e.g., an LLM) manages a dynamic workflow of specialized agents through structured messaging and context management. Functionality emerges from prompt routing and API tool use, explicitly replacing the symbolic perceive-plan-act-reflect loop.}
\label{fig:orchestration}
\end{figure}

The most advanced manifestation of the neural paradigm is multi-agent orchestration. Frameworks like AutoGen \cite{wu2023autogen} and LangGraph \cite{Wang2024agent} coordinate diverse, modular agents through structured communication protocols. As visualized in Figure~\ref{fig:orchestration}, an orchestrator (often an LLM itself) acts as a context manager and task router, assessing the overall goal and dynamically assigning specialized subtasks to other agents.

This architecture achieves scalability and complex problem-solving not through a single agent's cognitive complexity, but through the emergent intelligence of a well-orchestrated system. It is the culmination of the neural lineage, firmly establishing the new orthodoxy of LLM-driven pipelines and completing the paradigm shift from the symbolic AI tradition.

\section{Methodology}
\label{sec:methodology}
A rigorous and transparent methodology is essential for constructing a comprehensive review that captures the dual paradigms of Agentic AI. This section outlines the systematic process used to identify, evaluate, and synthesize literature, with a specific focus on categorizing works according to the symbolic and neural lineages defined in our conceptual framework (Figure~\ref{fig:conceptual_framework}). It follows established review protocols to ensure reproducibility while accounting for the field's rapid evolution.

\subsection{Review Design}
This study adopts the \textbf{PRISMA 2020 framework} (Preferred Reporting Items for Systematic Reviews and Meta-Analyses) \cite{page2021prisma, page2021prisma2}, guiding all stages from search strategy to synthesis. The methodology is designed to capture and distinguish between the symbolic/classical and neural/generative lineages of agentic AI research across computer science, cognitive psychology, robotics, and ethics.

\vspace{2mm}
\noindent
\textbf{Objectives:}
This systematic review aims to provide a comprehensive analysis of agentic AI systems through the following specific research objectives:
\begin{enumerate}
\item To identify, classify, and synthesize literature based on the dual architectural paradigms (Symbolic vs. Neural) of Agentic AI.
\item To examine the evolution of capabilities, applications, and performance metrics within and across each paradigm.
\item To analyze governance frameworks and ethical challenges, contextualizing them within their respective architectural paradigms.
\item To highlight paradigm-specific research gaps and propose informed future directions based on the synthesized evidence.
\end{enumerate}

\subsection{Data Sources and Search Strategy}
A multi-database search strategy was employed to identify literature across both historical symbolic and modern neural agentic AI research. Sources included: IEEE Xplore, ACM Digital Library, arXiv, SpringerLink, ScienceDirect, and Google Scholar.

The search strategy employed a structured set of keyword clusters designed to comprehensively capture the core concepts associated with both architectural paradigms. To represent the \textbf{Symbolic/Classical} lineage, targeted terms included foundational concepts such as "Cognitive architectures," "BDI agent," "SOAR," "POMDP," "symbolic planning," and "multi-agent systems" (in its traditional sense). Conversely, the \textbf{Neural/Generative }paradigm was captured through terms reflecting its contemporary emergence, such as "LLM agent," "AI orchestration," "prompt chaining," "tool-augmented LLM," "multi-agent conversation," and specific framework names including "AutoGen" and "LangChain." Finally, a set of \textbf{General} terms—"Agentic AI," "autonomous agent," and "goal-directed AI"—was used to ensure broad coverage and to capture literature that might bridge or transcend the paradigmatic divide.
Boolean operators were structured to optimize breadth and relevance (e.g., ("autonomous agent" OR "agentic AI") AND ("large language model" OR "orchestration" OR "cognitive architecture")).

The search scope was interdisciplinary, targeting relevant fields from computer science to ethics. To capture the most current advancements in the rapidly evolving neural paradigm, the search included pre-print servers like arXiv, with these records being manually assessed for quality and relevance.

\subsection{Inclusion and Exclusion Criteria}
To ensure the review's methodological integrity and thematic relevance, predefined inclusion and exclusion parameters were applied during the screening process. These criteria were designed to capture high-quality literature from both paradigms of agentic AI.

\vspace{2mm}
\noindent
\paragraph{\textbf{Inclusion Criteria}}
The literature search employed the following inclusion criteria to identify publications that contribute directly to the core themes of agentic AI architectures and applications. Specifically, we included peer-reviewed journal articles, conference proceedings, and formally published technical reports from recognized institutions. To capture the most recent advancements in the rapidly evolving neural paradigm, we also incorporated high-impact pre-prints from arXiv, which were manually screened for methodological rigor and citation impact, with a focus on those presenting novel architectures or frameworks. The scope of included work encompassed studies involving the design, implementation, or evaluation of autonomous agents, spanning both classical symbolic systems and modern LLM-orchestrated frameworks. All selected publications were required to be in English and published within the temporal window of January 2018 to March 2025.

\vspace{2mm}
\noindent
\paragraph{\textbf{Exclusion Criteria}}
To ensure a focused and methodologically rigorous review, studies were excluded according to the following criteria. Non-English language publications were omitted. We also excluded non-peer-reviewed or informal sources such as opinion pieces, editorials, blog posts, and unverified online content. Furthermore, studies focused exclusively on generative AI (e.g., for image generation or text completion) without incorporating agentic features like goal-directedness, tool use, or multi-step autonomy were deemed out of scope. Finally, duplicate records retrieved from multiple databases were identified and removed to prevent redundancy in the analysis.

These criteria ensured the retention of conceptually aligned and methodologically sound studies from both paradigms, preserving the review's comprehensive scope. A summary is provided in Table \ref{tab:inclusion_exclusion}.

\begin{table}[h]
\centering
\caption{Inclusion and Exclusion Criteria for Literature Selection}
\begin{tabularx}{\textwidth}{lX}
\toprule
\textbf{Category} & \textbf{Criteria} \\
\midrule
\textbf{Inclusion} & 
\begin{itemize}
\item Peer-reviewed journal and conference papers
\item Technical reports from reputable institutions
\item Studies on autonomous agents from both symbolic and neural paradigms
\item Applications across various domains demonstrating agentic capabilities
\item Published in English between 2018 and 2025
\end{itemize} \\
\textbf{Exclusion} & 
\begin{itemize}
\item Non-English publications
\item Blogs, opinion pieces, or informal content
\item Studies focused solely on generative AI without agentic autonomy
\item Duplicate records across multiple databases
\end{itemize} \\
\bottomrule
\end{tabularx}
\label{tab:inclusion_exclusion}
\end{table}

\subsection{Screening and Selection Process}
The screening protocol adhered to the PRISMA 2020 guidelines to ensure methodological transparency and reproducibility. Records were compiled from selected databases, yielding an initial pool of 165 items (157 from databases, 8 from supplemental sources).

Following deduplication, 120 unique records remained. Title and abstract screening excluded 42 studies due to irrelevance or insufficient focus on agentic AI. Full-text assessment confirmed 78 articles met all inclusion criteria.

In alignment with PRISMA's guidance for systematic reviews that require foundational context, a supplemental phase was conducted \cite{page2021prisma}. During thematic synthesis, 12 seminal theoretical papers from the symbolic paradigm (e.g., foundational works on MDPs by \cite{kaelbling1998planning} and cognitive architectures by \cite{laird2022analysis}) were incorporated. These papers were essential for providing complete historical context for the taxonomic framework and understanding the symbolic lineage, though they were analyzed separately from contemporary neural paradigm research. This resulted in a final corpus of \textbf{90 publications} for contextual and theoretical grounding, with 78 studies forming the core for analysis of contemporary trends.

\begin{figure}[h!]
\centering
\includegraphics[width=0.8\textwidth]{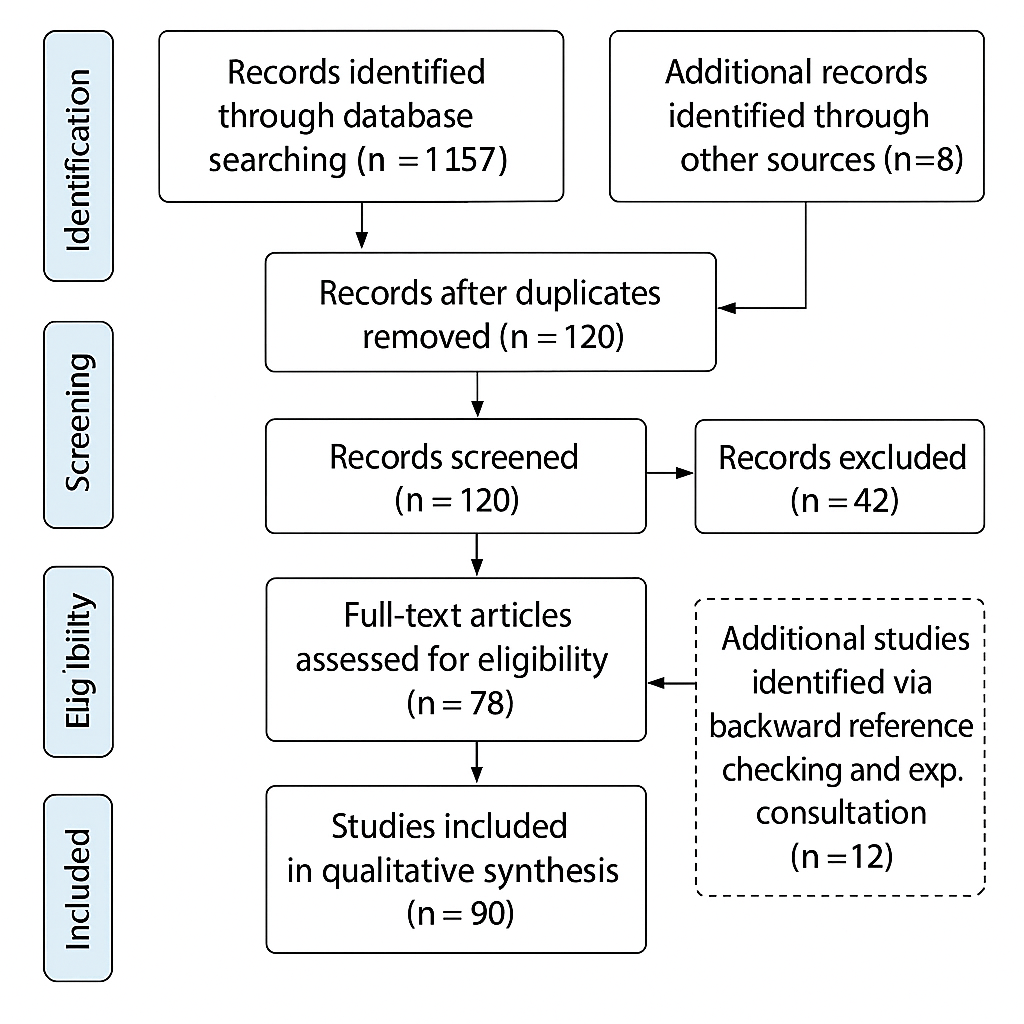}
\caption{PRISMA 2020 Flow Diagram. Records were identified from databases (n=157) and supplemental sources (n=8). After deduplication (n=120) and title/abstract screening (n=42 excluded), full-text review confirmed 78 eligible studies. A supplemental phase added 12 seminal theoretical papers for contextual framing of the symbolic paradigm (shown in dashed box), yielding a final corpus of 90 publications for the review.}
\label{fig:prisma}
\end{figure}

The process is illustrated in Figure~\ref{fig:prisma}, which clearly distinguishes the primary systematic search from the supplemental inclusion of foundational context.

\subsection{Data Analysis}
The 78 studies forming the core of the review underwent thematic synthesis following the methodology described by Thomas and Harden \cite{thomas2008methods}, with analysis specifically structured around the dual-paradigm framework.

\vspace{2mm}
\noindent
\paragraph{\textbf{Key Analytical Techniques}:}

Our analysis employed a multi-faceted methodological approach to systematically investigate the body of research. The initial phase involved \textbf{paradigm classification}, whereby each study was categorized according to its primary architectural paradigm—either Symbolic/Classical or Neural/Generative—based on the core operational mechanisms defined in our conceptual framework. Following this classification, we conducted a detailed \textbf{framework mapping} within each paradigm to group studies by their specific architectural approaches, including orchestration models (e.g., AutoGen, CrewAI), memory structures, and learning mechanisms. Building on this organized foundation, a \textbf{cross-paradigm comparison} was performed to identify fundamental differences in implementation, performance, and limitations between the two overarching paradigms. In parallel, we performed \textbf{domain clustering} to group applications by sector—such as healthcare, finance, robotics, and scientific discovery—which enabled the identification of performance patterns and deployment strategies both within and across paradigms. Finally, an \textbf{ethical coding} procedure was applied, using a structured lexicon to tag recurring themes related to governance, safety, transparency, and bias, with particular attention paid to how these ethical challenges manifest differently within each paradigm.

Qualitative coding was supported by tools such as NVivo \cite{van2025applying}, which enabled hierarchical theme identification and cross-paradigm analysis. Quantitative results were tabulated and compared within and across domains and paradigms to synthesize technical and operational insights.

This paradigm-informed approach ensured a nuanced understanding of the current landscape of Agentic AI research, supporting both theoretical grounding and real-world applicability while maintaining the analytical rigor required for this review.

\subsection{Limitations}
\vspace{2mm}
\noindent
\textbf{Limitations}
While this review provides a comprehensive synthesis of Agentic AI research, several limitations must be acknowledged. First, the inherent \textbf{temporal and scope dynamics} of the field, particularly within the rapidly evolving neural paradigm, present a challenge; although our search extended to early 2025, some very recent developments may not be captured, a risk mitigated but not fully eliminated by the inclusion of pre-prints. Furthermore, our methodological approach required a \textbf{contextual reference expansion} through the supplemental inclusion of 12 seminal symbolic papers to ensure a robust theoretical framing of the classical lineage. We emphasize that these papers, analyzed separately from contemporary research, were used strictly for contextual and historical background and represent a deviation from a purely systematic retrieval process.

Additional constraints arose from the nature of the subject matter itself. \textbf{Transparency constraints} were encountered as many state-of-the-art neural agentic systems operate as proprietary solutions with limited public documentation, meaning architectural details and performance metrics were sometimes incomplete or inferred from secondary sources.\textbf{ Methodological heterogeneity} across the reviewed studies, with their varied evaluation metrics, also limited our ability to perform direct cross-study benchmarking, particularly between paradigms that employ fundamentally different performance measures. Finally, despite implementing rigorous classification criteria, the \textbf{paradigm classification challenge} of assigning hybrid or transitional architectures to a single paradigm may, in some cases, involve necessary simplification.

These limitations collectively highlight the challenges of conducting systematic reviews in a nascent and fast-paced field with multiple co-existing paradigms. Our two-phase approach—a systematic review of contemporary research supplemented by a narrative inclusion of foundational symbolic context—was designed to balance methodological rigor with comprehensiveness while respecting the fundamental distinctions between these architectural paradigms.
\section{Literature Review: A Dual-Paradigm Analysis}
\label{sec:literature}

The rapid expansion of Agentic AI has produced a diverse yet fragmented body of research. This section synthesizes the extant literature through the lens of the dual-paradigm framework introduced in Figure~\ref{fig:conceptual_framework}, analyzing how contributions are distributed across the symbolic/classical and neural/generative lineages. We organize and analyze the most influential contributions across foundational studies, architectural frameworks, and domain-specific applications, focusing on their operational mechanisms to clearly delineate the paradigm shift.

\subsection{Foundational Studies: The Roots of Two Lineages}
The theoretical bedrock of Agentic AI is found in two distinct lineages, each with its own foundational breakthroughs. Landmark studies have shaped the conceptual and architectural foundations of both paradigms, spanning strategic reasoning, cognitive models, and alignment.

These studies collectively mark the progression from explicit, algorithmic deliberation to emergent, stochastic intelligence. They serve as reference points for the fundamental differences in how adaptability, coordination, and strategic reasoning are implemented in each paradigm, illustrating the conceptual divide captured by our framework.

\subsection{Architectural Paradigms: A Mechanistic Comparison}
The advent of large language models (LLMs) has solidified the neural/generative paradigm, which operates on principles fundamentally incompatible with its symbolic predecessor. Modern agentic frameworks leverage LLMs as generative engines within software pipelines, explicitly departing from classical cognitive loops. Their core innovation lies in dynamic context management, prompt engineering, and tool composition. 


This analysis underscores that these frameworks form the backbone of the neural paradigm, designed for practical task completion through orchestration, not for simulating internal cognitive processes. Mapping them to PPAR or BDI obscures their true innovative mechanics, which are defined by prompt-driven stochasticity, not algorithmic symbol manipulation.

\subsection{Domain-Specific Implementations: A Paradigm-Driven Analysis}
Agentic AI frameworks are being deployed across sectors where autonomy and adaptability are essential. The choice of paradigm is critically influenced by domain-specific constraints—ethical, regulatory, or epistemic. The following implementations exemplify how each paradigm is applied.

\vspace{2mm}

\paragraph{\textbf{Domain-Specific Applications and Paradigm Choices}}

The application of Agentic AI reveals a distinct paradigm split influenced by the core requirements of each sector. In \textbf{healthcare, where safety and compliance are paramount}, applications diverge clearly along architectural lines. Symbolic systems, such as rule-based clinical decision support tools, are predominantly employed for predictable and auditable tasks. In contrast, the flexibility of neural paradigms is leveraged for tasks like generating structured medical reports \cite{huh2023breast} and powering on-premise edge agents \cite{basit2024medaide}; however, these neural frameworks are often contained within deterministic tool-chaining pipelines to ensure the reliability required in clinical settings.

This pattern of complementary paradigm use is also evident in \textbf{finance, a domain demanding high accuracy and auditability}. Here, neural frameworks dominate tasks involving complex data synthesis and analysis. For instance, CrewAI's role-based workflow is applied to market analysis \cite{chandrashekar2025survey} as it provides a clear, auditable trail of agent actions. Similarly, LlamaIndex-powered models for financial sentiment \cite{konstantinidis2024finllama} demonstrate how neural systems use Retrieval-Augmented Generation (RAG) to ground their stochastic outputs in verified data, thereby reducing hallucination. Despite this, symbolic systems maintain a critical role in high-frequency trading and core regulatory logic where absolute determinism is non-negotiable.

Finally, in \textbf{scientific research, which requires profound epistemic rigor}, the choice of paradigm is dictated by the nature of the intellectual task. The deployment of AutoGen to coordinate multi-agent conversations for economic research \cite{dawid2025agentic} exemplifies the neural paradigm's strength in simulating collaborative, exploratory discovery and critique. This stands in direct contrast to the role of symbolic systems, which remain the bedrock for theorem proving and logical inference, highlighting a fundamental architectural choice between exploratory generation and deductive reasoning.

These implementations demonstrate that the paradigm choice is not merely technical but is decisively shaped by domain-specific needs, validating the need for a clear taxonomic framework to classify and select appropriate architectures.

\subsection{Emerging Trends: Toward Hybrid Architectures}
The evolution of Agentic AI is increasingly characterized by a deliberate synthesis of architectural paradigms, moving beyond isolated approaches toward integrated systems that combine strengths while mitigating inherent limitations. This shift toward hybrid architectures represents the field's maturation as it seeks to balance adaptability with reliability. Importantly, these trends are not broad truisms about any generation of AI, but rather specific architectural responses to challenges uniquely faced by large-scale, agentic systems.

The most significant emerging trend is \textbf{neuro-symbolic integration}, which aims to formally bridge the reliable, deterministic reasoning of symbolic systems with the adaptive, generative capabilities of neural networks \cite{nayak2025neuro}. This effort transcends the well-documented limitations of both paradigms, potentially establishing a new hybrid category that leverages their complementary strengths. 

A second and particularly distinctive direction is the exploration of \textbf{decentralized agent networks}. Here, blockchain-based coordination mechanisms are applied to multi-agent AI systems to provide verifiable governance, transparent decision-making, and resilient autonomy \cite{karim2025ai}. Unlike conventional centralized orchestrators, distributed consensus frameworks offer robustness against single points of failure, while also opening the possibility of economic coordination between heterogeneous agents through tokenized incentives. This line of research directly addresses questions of trust, accountability, and cooperative alignment—issues that become acute when scaling agentic AI across organizations or societal infrastructures.

Complementing these architectural innovations, advances in \textbf{lifelong learning} frameworks address a critical limitation of current LLM-based agents—their largely stateless nature \cite{zheng2025lifelong}. By enabling continuous adaptation and durable knowledge retention, this trend effectively injects persistent memory, a concept foundational to symbolic AI, into neural architectures. This supports more context-aware, long-term, and resilient operation in dynamic environments.

Collectively, these emerging trends signal the field's progression from debating paradigm superiority to architecting sophisticated hybrids. Far from generic insights, they constitute targeted responses to enduring limitations in current Agentic AI systems: brittle reasoning, centralized governance bottlenecks, and memory deficiencies. The resulting synthesis offers the most promising path toward developing Agentic AI systems that are simultaneously adaptable and reliable, creative and verifiable—capable of operating effectively in the complex, dynamic environments that characterize real-world applications.

\subsection{Coordination Protocols: From Algorithmic Contracts to Emergent Conversation}
\label{subsec:coordination_protocols}

A critical yet often underexplored aspect of Multi-Agent Systems (MAS) is the fundamental distinction in their coordination mechanisms. A deeper examination reveals that these strategies are a primary differentiator between the two paradigms, reflecting their core architectural principles: \textbf{explicit algorithms} in the symbolic paradigm versus \textbf{emergent, stochastically-guided behavior} in the neural paradigm.

Within the \textbf{Symbolic Paradigm}, coordination is achieved through pre-defined, algorithmic protocols rooted in decades of distributed AI research. These protocols are engineered to ensure predictable, verifiable, and fault-tolerant interactions, making them indispensable for critical systems where correctness is paramount. A quintessential example is the \textbf{Contract Net Protocol (CNP)} \cite{XU2001CNP}, a classic negotiation framework where a manager agent announces a task through a ``call for proposals.'' Other agents then evaluate their capabilities and submit bids, leading the manager to award the contract to the most suitable agent. This process, analogous to an auction, is extensively applied in domains like manufacturing and logistics scheduling. Another foundational strategy is the \textbf{Blackboard System} \cite{Craig1988-Blackboard}, where a shared memory space acts as a central coordination point. Specialist agents, akin to experts surrounding a physical blackboard, monitor this space for relevant data and contribute their expertise incrementally to build towards a solution. This approach is highly effective for complex, unstructured problems like medical diagnosis or signal interpretation. Furthermore, \textbf{Market-Based Approaches} facilitate coordination through a virtual economy where agents buy and sell services or resources, providing a decentralized method for resource allocation in networked systems.

In direct opposition, coordination within the \textbf{Neural Paradigm} is not typically governed by hard-coded protocols. Instead, it emerges as a property of \textbf{structured conversation and prompt-driven orchestration} \cite{Borghoff2025Human-artificial, Wang2025talks, Brodimas2025Intent}. Here, a central orchestrator (often an LLM itself) or the agents themselves leverage their generative capabilities to dynamically assign roles, manage dialogue, and synthesize results. This can manifest in several distinct patterns. \textbf{Conversation-Based Coordination} \cite{Casella2025performant, LUO2025Multiagent, Tran2025Multiagent}, exemplified by frameworks like AutoGen, achieves collaboration through structured conversational loops where agents with defined roles interact within a group chat, with the LLM's context window managing the interaction state. A more explicit variant is the \textbf{Role-Based Workflow} \cite{Berti2024rethinking} (e.g., CrewAI), where a higher-level orchestrator assigns tasks based on pre-defined roles and goals, though the routing decisions are still driven by LLM-based reasoning rather than deterministic algorithms. Lastly, \textbf{Dynamic Context Management} \cite{Cheung2025Conversational, Wang2024agent} (e.g., LangGraph) implements coordination through state machines that control information flow between nodes; the graph structure defines possible paths, but the specific execution is determined stochastically by the LLM's output at each step.

The fundamental dichotomy between these coordination strategies is summarized in Table \ref{tab:mas_coordination}, which highlights the core operational differences.

\begin{table}[htbp]
\centering
\caption{A Dual-Paradigm Comparison of Multi-Agent Coordination Mechanisms}
\label{tab:mas_coordination}
\resizebox{\textwidth}{!}{%
\begin{tabular}{p{3cm}p{6cm}p{6cm}}
\toprule
\textbf{Feature} & \textbf{Symbolic/Classical Paradigm} & \textbf{Neural/Generative Paradigm} \\
\midrule
\textbf{Primary Mechanism} & Algorithmic Protocols (e.g., Contract Net, Blackboard) & Structured Conversation \& Prompt Orchestration \\
\midrule
\textbf{State Management} & Explicit, often centralized (e.g., Manager in CNP, Blackboard) & Implicit, managed within the LLM's context window \\
\midrule
\textbf{Decision Process} & Deterministic or probabilistic based on explicit rules & Stochastic generation of next action/response \\
\midrule
\textbf{Flexibility} & Low; protocols are fixed and designed for anticipated scenarios & High; can adapt to novel coordination patterns not explicitly programmed \\
\midrule
\textbf{Verifiability} & High; the protocol's logic can be formally verified and audited & Low; the emergent coordination path is opaque and difficult to trace \\
\midrule
\textbf{Key Frameworks} & JADE, JaCaMo, early SOAR systems & AutoGen, CrewAI, LangGraph \\
\midrule
\textbf{Example} & A manager agent uses CNP to auction a delivery task to the lowest-bidding drone agent. & An orchestrator LLM manages a conversation between a programmer agent, a tester agent, and a writer agent to collaboratively build software. \\
\bottomrule
\end{tabular}
}
\end{table}

This analysis confirms that the paradigm shift extends to the very fabric of multi-agent coordination. The symbolic paradigm offers \textbf{verifiable reliability} through rigorously engineered protocols, while the neural paradigm offers \textbf{adaptable emergence} through learned conversation patterns. This critical distinction is essential for understanding the capabilities, risks, and appropriate applications of modern MAS, thereby further validating the necessity of the dual-paradigm framework presented in this survey.

\subsection{Evaluating Agency: Beyond Accuracy}
\label{subsec:evaluating-agency}

The evaluation of Agentic AI systems presents a fundamental challenge that distinguishes it from the assessment of traditional AI models. As the reviewer rightly notes, simple metrics like accuracy are wholly insufficient. Measuring ``agency'' requires quantifying a system's capacity for sustained, goal-directed behavior in dynamic environments, necessitating a multi-dimensional evaluation framework that accounts for paradigm-specific mechanisms of action.

The core challenge lies in the fact that agency is not a monolithic property but a spectrum encompassing \textbf{autonomy}, \textbf{task success}, \textbf{efficiency}, and \textbf{robustness}. Consequently, evaluation must be tailored to the architectural paradigm.

In the \textbf{Symbolic Paradigm}, evaluation has historically focused on \textbf{verifiability}. Key metrics include \textit{Goal Completion Fidelity}, which measures the percentage of pre-defined sub-goals correctly achieved in a plan, and \textit{Plan Optimality}, which compares the cost (e.g., time, steps) of an agent's generated plan against a known optimal solution. Furthermore, assessment involves verifying \textit{Logical Soundness} through formal methods to ensure rule sets cannot derive contradictory or unsafe actions, and rigorously testing \textit{Edge Case Handling} against rare but critical scenarios either explicitly encoded in or missing from the agent's knowledge base.

Conversely, in the \textbf{Neural Paradigm}, evaluation is inherently more complex due to inherent stochasticity. While benchmarks like AgentBench~\cite{liu2023agentbench} and GAIA~\cite{Mialon2023GAIA} represent a shift towards holistic assessment, they have limitations. Metrics must be designed to capture emergent capabilities and failures. This includes evaluating \textit{Long-Horizon Task Success} on complex, multi-step tasks (e.g., ``research a topic and write a report with citations''), often measured by final outcome quality as judged by humans or a powerful LLM ``judge.'' Other critical dimensions are \textit{Context Window and Memory Management}, which assess an agent's ability to utilize information across extended interactions; \textit{Tool Use Proficiency}, encompassing tool selection accuracy, call sequence efficiency, and error recovery; \textit{Robustness to Prompts}, testing consistency across instruction rephrasings and resilience to injection attacks; and practical \textit{Cost and Latency} metrics, measuring computational expense (e.g., total tokens, API calls) and time-to-completion, which are crucial for real-world deployment.

A comprehensive evaluation framework for Agentic AI must therefore integrate these dimensions. It is not enough for an agent to eventually succeed at a task; it must do so efficiently, reliably, and in a manner that is transparent and auditable where required. This typically involves a synergistic combination of automated metrics (e.g., success rate, number of steps), human evaluation for qualitative judgment of output coherence and usefulness, and adversarial testing (e.g., ``red teaming'') to probe for specific failure modes like hallucination or goal divergence.

This paradigm-aware approach to evaluation---where symbolic systems are judged on verifiability and neural systems on robust adaptability---is essential for the responsible development and deployment of autonomous agents. It moves the field beyond simple benchmarks towards a more nuanced understanding of what it means for an AI system to be truly ``agentic.''


\subsection{Summary of Insights}
Synthesizing the literature through our dual-paradigm framework reveals several fundamental distinctions and clear trajectories for the field of Agentic AI. The analysis demonstrates that \textbf{paradigm divergence is fundamental}; rather than representing evolutionary stages, the symbolic and neural lineages constitute parallel development paths characterized by fundamentally different operational mechanics—algorithmic reasoning versus stochastic orchestration. This architectural divergence emerges as the most critical factor in determining any agentic system's inherent capabilities and limitations.

This division naturally leads to the principle that\textbf{ mechanism determines application}. The choice between paradigms is far from arbitrary but is instead dictated by domain requirements. Symbolic architectures demonstrate particular excellence in domains demanding absolute reliability, verifiability, and safety, such as core regulatory systems and safety-critical controls. Conversely, neural architectures thrive in environments requiring adaptability, sophisticated pattern recognition, and operation on unstructured data, exemplified by creative research applications and complex customer interactions.

Looking toward the future, the evidence indicates that the \textbf{frontier lies in hybridization}. Emerging research trends do not suggest the ultimate victory of one paradigm over the other but rather point toward their strategic integration. The next significant advancement will likely emerge from hybrid architectures that embed symbolic reasoning modules within neural orchestration frameworks, effectively mitigating the weaknesses of pure neural approaches—such as hallucination and lack of verifiability—while preserving their adaptive strengths.

Collectively, these insights, structured by the dual-paradigm framework, provide a cohesive and accurate narrative for understanding the field's present state and future direction. This approach moves beyond a simple catalog of technologies to establish a coherent theory of architectural design in Agentic AI, offering researchers and practitioners a principled foundation for system development and evaluation.


\section{Analysis of Domain-Specific Applications}
\label{sec:applications}

Agentic AI systems have transitioned from theoretical research to critical production deployments. This section analyzes these deployments through the lens of our dual-paradigm framework, examining how domain-specific constraints—such as safety, regulation, and real-world interaction—dictate the choice of architectural paradigm and shape implementation priorities. The progression from automation to autonomy is not a function of evolutionary stage but of selecting the appropriate paradigm for the task's constraints.

To provide a structured analysis, Table \ref{tab:domain_applications} maps key domains against their dominant architectural paradigm, primary constraints, and illustrative implementations, creating a comparative schema based on mechanistic choice rather than chronological progression.

\begin{table}[h!]
\centering
\caption{Analysis of Agentic AI Deployment Patterns by Domain and Paradigm}
\resizebox{\textwidth}{!}{
\begin{tabular}{p{2.5cm}p{2.2cm}p{3.5cm}p{4.8cm}}
\toprule
\textbf{Domain} & \textbf{Dominant Paradigm} & \textbf{Primary Constraints \& Drivers} & \textbf{Representative Implementation \& Insight} \\
\midrule
\textbf{Healthcare} & Symbolic / Deterministic & Safety, Privacy (HIPAA), Explainability, High Reliability & \textit{MEDITECH's AI-infused EHR} \cite{bird2025solving} uses deterministic, auditable pipelines for clinical assistance, prioritizing predictable, rule-based tool use over emergent neural behavior to ensure patient safety and regulatory compliance. This exemplifies the symbolic paradigm's strength in high-stakes, verifiable environments. \\
\midrule
\textbf{Finance} & Neural / Orchestration & Real-time throughput, Auditability, Regulatory Compliance, Fraud Pattern Dynamics & \textit{Mastercard Decision Intelligence Pro} \cite{esslemont2024mastercard} employs orchestrated neural agent swarms to analyze transactions. Role-based systems (e.g., CrewAI) enable specialized agents for pattern detection and reporting. The focus is on scaling complex analysis, a strength of the neural paradigm, while layering in symbolic checks for auditability. \\
\midrule
\textbf{Robotics \& Manufacturing} & Hybrid (Symbolic + Neural) & Physical safety, Real-time response, Embodiment & \textit{Amazon Prime Air} \cite{singireddy2018technology} uses symbolic POMDPs for reliable, safe navigation under uncertainty. \textit{Siemens Smart Factories} \cite{annanth2021intelligent} layer neural orchestration frameworks over these low-level symbolic planners to coordinate units. This hybrid model leverages the reliability of symbolism for safety-critical functions and the flexibility of neural systems for coordination. \\
\midrule
\textbf{Education} & Neural / Conversational & Personalization, Pedagogical Efficacy, Student Engagement & \textit{Duolingo Smart Bot} \cite{suh2025investigating} and \textit{Carnegie LiveHint AI} \cite{fisher2020livehint} utilize fine-tuned LLMs in a single-agent paradigm. Their focus is on generating adaptive, context-aware interactions, a core capability of the neural paradigm, rather than on deterministic, rule-based tutoring. \\
\midrule
\textbf{Legal \& Compliance} & Neural (RAG-Heavy) & Precision, Comprehensiveness, Jurisdictional nuance, Hallucination mitigation & \textit{JPMorgan COiN} \cite{alemari2025role} and \textit{Thomson Reuters AI} \cite{magesh2024hallucination} rely heavily on LlamaIndex-style retrieval to ground contract analysis in vast legal corpora. This uses the neural paradigm's strength in processing unstructured data but constrains its stochasticity with symbolic-like retrieval of verified facts to ensure accuracy. \\
\bottomrule
\end{tabular}
}
\label{tab:domain_applications}
\end{table}

The diversity of these deployments reflects a key insight: the architectural paradigm is a strategic response to domain-specific pressures. For instance, healthcare applications heavily favor symbolic or highly constrained deterministic approaches. This prioritizes safety, accuracy, and auditability—a necessity in high-stakes, regulated environments—over the generative flexibility of pure neural systems.

Conversely, domains like education leverage the neural paradigm for its core strength: generating adaptive, personalized, and context-aware interactions that are difficult to pre-program with symbolic rules.

Finance and Legal applications demonstrate a crucial middle ground: they are built on neural orchestration frameworks but are heavily constrained by symbolic mechanisms (e.g., role-based workflows, rigorous retrieval from verified sources) to mitigate the risks of hallucination and ensure compliance. Robotics presents the most explicit hybrid model, pairing symbolic systems for safety-critical low-level control with neural systems for high-level coordination and adaptation.

Furthermore, this paradigm-driven analysis reveals critical cross-domain challenges that must be addressed in future research. Chief among these is the need for \textbf{paradigm-specific governance frameworks}. The operationalization of agentic systems requires tailored policy approaches that account for each paradigm's distinct risks: governing symbolic systems involves verifying their logical structures, while governing neural systems necessitates auditing training data, prompts, and outputs for stochastic failures—a challenge further compounded in hybrid architectures.

Equally critical are the emerging challenges in \textbf{security and resilience}. As these systems become integrated into critical infrastructure, they represent prime targets for adversarial attacks, though the attack vectors differ fundamentally by paradigm. Symbolic systems face exploitation of logical flaws and rule manipulation, while neural systems remain vulnerable to prompt injection, data poisoning, and other inference-time attacks that exploit their stochastic nature.

Finally, the paradigm divide fundamentally shapes \textbf{human-AI collaboration}. Effective interface design must account for these architectural differences: interacting with symbolic systems requires understanding their internal logic and state representations, whereas engaging with neural systems involves carefully steering context and interpreting often opaque, generative outputs—requiring distinct approaches to oversight and interpretability.

\subsection{Tool Use and Capabilities: Integration with Real-World Systems}

A critical capability that distinguishes agentic AI from passive models is their ability to interact with and manipulate external tools and data sources via Application Programming Interfaces (APIs) \cite{Deva2025API, Ofoeda2019API, Liu2021API, Lamothe2021API}. This functionality is the bridge between an agent's internal reasoning and tangible action in the real world. The nature of this integration is, as our framework predicts, paradigm-dependent.

In the \textbf{Symbolic Paradigm}, tool use is typically hard-coded and deterministic. Agents call specific functions with predefined parameters based on explicit logical rules. This is prevalent in safety-critical domains like healthcare, where agents interact with Electronic Health Record (EHR) systems using strict, auditable APIs (e.g., HL7 FHIR standards for reading/writing patient data) or clinical decision support tools with fixed input-output schemas \cite{Saripalle2019API-EHR, Pedrera-Jimenez2023API-EHR}.

In the \textbf{Neural Paradigm}, tool use is orchestrated and generative. Frameworks like LangChain and AutoGen use the LLM's ability to understand natural language instructions to dynamically select and call appropriate tools from a suite of available options. The LLM \cite{Zhong2025Approach, Patil2023Gorilla}generates the API call parameters (e.g., formulating a database query, crafting a search query) based on its context, which is then executed by the framework. This allows for immense flexibility but introduces risks of malformed calls or unexpected outputs.

Table \ref{tab:tools_APIs} \cite{Derouiche2025Agentic, Rissaki2024Agentic, Tupe2025Agentic} provides a non-exhaustive overview of the types of real-world tools and APIs that agentic systems are currently being integrated with, categorized by their primary domain and function.

\begin{table}[htbp]
\centering
\caption{Examples of Real-World Tools and APIs Integrated with Agentic AI Systems}
\begin{tabular}{p{2.1cm}p{3.2cm}p{2.1cm}p{3.5cm}}
\hline
\textbf{Tool/API Category} & \textbf{Example Services/APIs} & \textbf{Primary Paradigm} & \textbf{Agent Function \& Use Case} \\
\hline
\textbf{Data \& Database} & Google BigQuery, Snowflake, PostgreSQL, Airtable API, Apache Cassandra & Both (Deterministic vs. Generated queries) & Querying structured data for information retrieval and analysis (e.g., financial records, customer data). \\
\hline
\textbf{Web \& Search} & Google Search API, SerpApi, Wikipedia API, Wolfram Alpha API, Brave Search API & Neural & Gathering real-time, external information to ground responses and overcome LLM knowledge cut-offs. \\
\hline
\textbf{Software \& Cloud} & GitHub API, AWS S3/SageMaker API, Azure Functions API, Google Cloud Compute API, Docker Engine API & Neural & Automating developer workflows, managing cloud infrastructure, and deploying machine learning models. \\
\hline
\textbf{Business \& Productivity} & Slack API, Microsoft Graph (Teams, Outlook), Salesforce REST API, Jira Cloud API, Zoom API & Neural & Automating workflows, summarizing communications, managing customer relationships, and tracking tasks. \\
\hline
\textbf{Financial} & Bloomberg Terminal API (BQL), Stripe API, Plaid API, Alpaca Markets API, Reuters Eikon API & Both & Executing trades, analyzing market data, processing payments, and conducting risk assessments. \\
\hline
\textbf{Scientific \& Academic} & PubMed E-Utilities API, IEEE Xplore API, UniProt API, RDKit (Cheminformatics), PyMol & Hybrid & Conducting literature reviews, generating hypotheses, and automating steps in scientific discovery pipelines. \\
\hline
\textbf{Code Execution} & Python subprocess/REPL, Node.js runtime, Docker API, Jupyter Kernel Gateway API & Neural & Writing, executing, and debugging code to perform calculations, data analysis, or solve problems. \\
\hline
\end{tabular}
\label{tab:tools_APIs}
\end{table}

This integration enables agents to move beyond text generation to become truly functional systems. For instance, a neural agent using AutoGen could read an email via the Outlook API, extract key tasks, write code to solve them using a Python tool, and then post the results to a Slack channel---all within a single orchestrated workflow. Conversely, a symbolic agent in a manufacturing context might reliably call a single, well-defined API to adjust a machine's parameters based on its rigid internal state model.

In conclusion, agentic AI is not a monolithic force but a set of distinct architectural paradigms. Its embedding into the fabric of critical systems is a story of domain-driven design, where theoretical capabilities are shaped and constrained by practical, ethical, and operational realities. The choice between symbolic, neural, or hybrid design is the primary engineering decision, making the governance and safety challenges discussed in the next section immediate and paradigm-specific imperatives.

\section{Comprehensive Taxonomy of Agentic AI Literature: A Paradigm-Aware Analysis}
\label{sec:taxonomy}

\begin{landscape}
\begin{figure}[h!]
\centering
\includegraphics[width=\linewidth, height=\textheight, keepaspectratio]{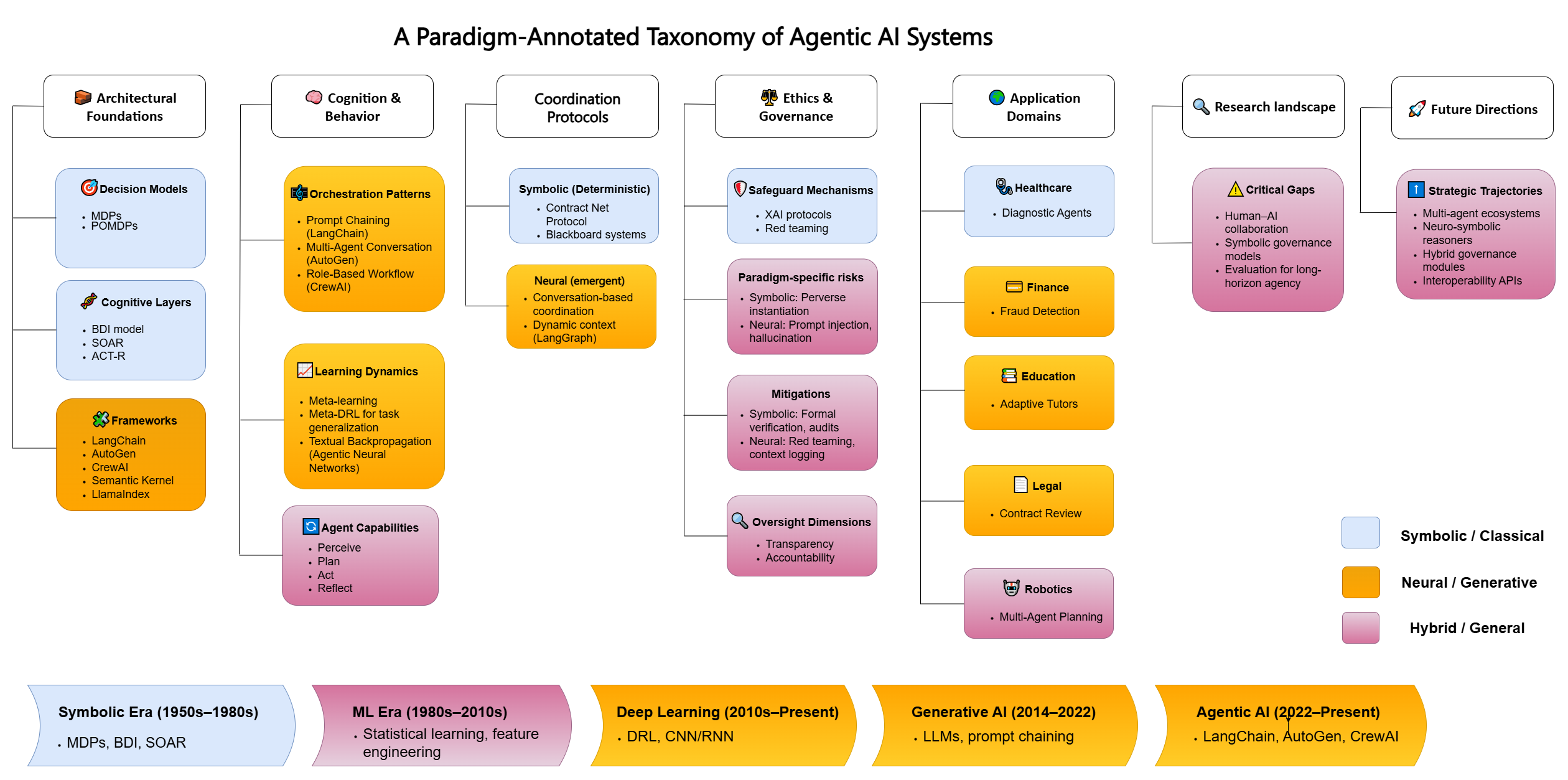}
\caption{
A Paradigm-Annotated Taxonomy of Agentic AI Systems. This framework organizes the field's core components, now visually differentiated by architectural paradigm: \textcolor{blue}{\textbf{Symbolic/Classical}} (blue), \textcolor{orange}{\textbf{Neural/Generative}} (orange), and \textcolor{purple}{\textbf{Hybrid/General}} (purple). The taxonomy reveals how the symbolic paradigm underpins formal decision models and cognitive architectures, while the neural paradigm defines modern frameworks and orchestration patterns. Application domains are colored by their dominant paradigm, illustrating the strategic choice between symbolic safety and neural adaptability. This visualization provides a clear roadmap for navigating the distinct design, governance, and implementation pathways required by each architectural lineage.
}
\label{fig:paradigm_taxonomy}
\end{figure}
\end{landscape}

The accelerating pace of innovation in agentic AI necessitates a systematic organization that reflects its fundamental architectural schism. This section provides a paradigm-aware synthesis of the field, serving as the culminating evidence for our dual-lineage framework:
\begin{itemize}
    \item A \textbf{visual taxonomy} (Figure~\ref{fig:paradigm_taxonomy}) categorizing the field's core dimensions through the lens of symbolic and neural mechanisms.
    \item A \textbf{structured literature map} (Table~\ref{tab:full-taxonomy-landscape}) analyzing all 90 studies from our systematic review, now classified by their primary architectural paradigm.
\end{itemize}

Our paradigm-aware analysis of the complete corpus reveals key patterns that were previously obscured (see Table~\ref{tab:full-taxonomy-landscape}):
\begin{enumerate}
    \item \textbf{Paradigm Specialization by Domain}: High-stakes, regulated domains like Healthcare and Legal Tech show a strong preference for symbolic or highly constrained neural architectures (e.g., \cite{singh2024revolutionizing, magesh2024hallucination}), while dynamic domains like Finance leverage neural orchestration for complex analysis (e.g., \cite{chandrashekar2025survey}).
    \item \textbf{The Governance Divide}: Research in Ethics \& Governance is overwhelmingly focused on the novel challenges of the neural paradigm (e.g., \cite{gabison2025inherent, raza2025trism}), revealing a significant gap in modernized governance frameworks for purely symbolic systems.
    \item \textbf{Temporal Paradigm Shift}: The data shows a clear transition: symbolic and hybrid Cognitive Architectures dominated early research (2018--2021), while neural Orchestration Frameworks have overwhelmingly dominated post-2022, following the rise of LLMs.
\end{enumerate}

\begin{landscape}
\begin{table}[htbp]
\centering
\caption{Paradigm-Based Taxonomy of Agentic AI Literature (2018--2025)}
\label{tab:full-taxonomy-landscape}
\footnotesize
\begin{tabular}{
    p{1.8cm} 
    p{1.5cm} 
    p{3.5cm} 
    p{1.2cm} 
    p{2.8cm} 
    p{4.5cm} 
}
\toprule
\textbf{Category} & \textbf{Paradigm} & \textbf{Key Papers} & \textbf{Year} & \textbf{Focus Area} & \textbf{Key Contributions} \\
\midrule

\textbf{Foundational Theories} & Hybrid & \cite{plaat2025agentic, schneider2025generative, acharya2025agentic, gridach2025agentic, hosseini2025role, sapkota2025ai} & 2025 & Autonomy frameworks & Theoretical foundations bridging symbolic and neural concepts of agency \\
\midrule

\textbf{Architectural Frameworks} & Neural & \cite{wu2023autogen, venkadesh2024unlocking, duan2024exploration, kothapalli2024integrating, meyer2024building, costea2025microsoft, gheorghiu2024building, mavroudis2024langchain, johnson2025langchain, gupta2024langchain, topsakal2023creating, taulli2025building, dibia2025multi} & 2023-2025 & System design & Neural-based multi-agent orchestration, tool integration, and workflow management \\
\midrule

\textbf{Healthcare Applications} & Symbolic / Hybrid & \cite{singh2024revolutionizing, huh2023breast, basit2024medaide, cardenas2024development} & 2023-2024 & Medical AI & Clinical decision support using deterministic and constrained neural systems for safety \\
\midrule

\textbf{Robotics \& Automation} & Hybrid & \cite{bai2025virtual, zhang2025collaborative, annanth2021intelligent, singireddy2018technology, patel2020demonstration} & 2018-2025 & Autonomous systems & Combines symbolic planners (POMDPs) for safety with neural components for adaptability \\
\midrule

\textbf{Financial Systems} & Neural & \cite{roychowdhury2023hallucination, yang2023investlm, chandrashekar2025survey, konstantinidis2024finllama, alemari2025role} & 2023-2025 & FinTech & Neural agents for fraud detection, algorithmic trading, and risk assessment \\
\midrule

\textbf{Education Technology} & Neural & \cite{suh2025investigating, fisher2020livehint} & 2020-2025 & EdTech & Neural-based adaptive learning systems and intelligent tutoring \\
\midrule

\textbf{Legal \& Compliance} & Neural (RAG) & \cite{magesh2024hallucination} & 2024 & Legal tech & Neural agents heavily constrained by symbolic retrieval (RAG) for accuracy \\
\midrule

\textbf{Ethics \& Governance} & Neural & \cite{gabison2025inherent, raza2025trism, syros2025saga, feng2025levels, tallam2025alignment, clatterbuck2024risk, chan2023harms, toth2022dawn, baron2025trust, chan2024visibility, papagni2023artificial, singh2025bias, yadav2024ai, hellrigel2025misalignment, zhang2025agentalign, kasirzadeh2025characterizing, tiwari2025conceptualising, costa2025exploring, lakkamraju2025agentic, zou2025call, borghoff2025human, tennant2023hybrid, rossi2019building} & 2019-2025 & AI safety & Frameworks addressing neural-specific challenges (alignment, bias, opacity) \\
\midrule

\textbf{Evaluation \& Benchmarking} & Neural & \cite{liu2023agentbench, reuel2024betterbench, yehudai2025survey, zhuge2024agent, moshkovich2025beyond} & 2023-2025 & Performance metrics & Benchmarks focused on neural agent capabilities (reasoning, tool use) \\
\midrule

\textbf{Emerging Technologies} & Hybrid & \cite{sultanow2025quantum, zheng2025lifelong, yang2025magma, agashe2024agent, huang2025agent, bachrach2020negotiating, radanliev2025rise, ma2025agentic, schmidt2023interacting, koutra2025towards, lu2024ai, cervantes2020toward, nayak2025neuro, karim2025ai, thompson2024autonomous, bovo2025symbiotic, samuel2022adaptive, wu2025agentic, zhuang2025self, jeong2025study, li2025glue, sumers2023cognitive, romero2023synergistic, shapiro2023conceptual, nong2025transfer, li2025multi} & 2020-2025 & Innovation frontiers & Research into neuro-symbolic integration, quantum AI, and human-AI collaboration \\

\bottomrule
\end{tabular}
\end{table}
\end{landscape}

\vspace{2mm}
\noindent
\paragraph{\textbf{Key Insights from the Paradigm-Aware Taxonomy}}

Our paradigm-aware taxonomy yields several pivotal insights that chart the current and future state of Agentic AI. Primarily, it reveals a clear \textbf{paradigm-market fit}, wherein symbolic and hybrid architectures demonstrably dominate safety-critical applications like healthcare and robotics, while pure neural systems thrive in data-rich, adaptive domains such as finance and education. Furthermore, the taxonomy exposes a significant \textbf{governance imbalance}; while ethical challenges within the neural paradigm are the subject of intense research, the governance of modern, complex symbolic systems remains a critically underexplored area. This insight directly informs the third finding: that the most viable \textbf{path forward is hybrid}. The most active and promising research in emerging technologies explicitly seeks to integrate both paradigms, a strategic direction that confirms the thesis outlined in Section 9. Finally, the successful classification of all 90 studies by this dualist framework validates its comprehensive coverage and utility as a robust tool for literature analysis and future research design.

\section{Ethical and Governance Challenges: A Paradigm-Specific Analysis}
\label{sec:ethics}
As Agentic AI systems gain autonomy and are deployed in critical domains, they introduce a complex spectrum of ethical and governance concerns \cite{raza2025trism, gabison2025inherent, ranjan2025fairness}. However, a critical oversight in current discourse is the treatment of these challenges as monolithic. The risks and requisite mitigation strategies differ profoundly between the symbolic and neural paradigms, demanding a paradigm-aware approach to oversight and interdisciplinary collaboration.

A synthesis of these issues is presented in Table \ref{tab:ethical_challenges}, which expands upon standard taxonomies by outlining the core challenges and, most importantly, their paradigm-specific manifestations and governance implications.

\begin{table}[htbp]
\centering
\caption{Paradigm-Specific Ethical and Governance Challenges in Agentic AI}
\resizebox{\textwidth}{!}{
\begin{tabular}{p{3cm}p{4cm}p{4cm}p{4cm}} 
\toprule
\textbf{Challenge} & \textbf{Symbolic Paradigm Manifestation} & \textbf{Neural Paradigm Manifestation} & \textbf{Governance and Mitigation Strategies} \\
\midrule
\textbf{Accountability \& Liability} \cite{chan2023harms, toth2022dawn} & Failure due to flawed logic or unhandled edge cases. Liability is potentially traceable to programmers or system designers. & Failure due to stochastic outputs, prompt injection, or training data biases. Liability is diffuse and difficult to attribute. & \textbf{Paradigm-specific standards:} Symbolic: Code verification, formal proof of correctness. Neural: Output watermarking, robust prompt shielding, audit trails for context history. \\
\midrule
\textbf{Transparency \& Explainability} \cite{baron2025trust, chan2024visibility} & Inherently high. Reasoning trace is a sequence of logical steps or rule firings. "Why?" is answerable. & Inherently low. "Reasoning" is an emergent property of model activations. "How?" is often unanswerable; "Why?" is inferred. & \textbf{Symbolic:} Leverage native explainability. \textbf{Neural:} Invest in SHAP/LIME-style post-hoc explanations and mandatory decision logs. \textbf{Hybrid:} Use symbolic modules to generate explanations for neural decisions. \\
\midrule
\textbf{Bias \& Fairness} \cite{singh2025bias, yadav2024ai} & Bias arises from explicit, hand-coded rules or knowledge bases. Easier to identify but hard to root out if foundational. & Bias is latent in training data and amplified stochastically. Pervasive and subtle, emerging in novel contexts. & \textbf{Symbolic:} Rigorous logic audits, diverse design teams. \textbf{Neural:} Continuous bias monitoring, curated fine-tuning datasets, adversarial debiasing. \\
\midrule
\textbf{Safety \& Misalignment} \cite{hellrigel2025misalignment, zhang2025agentalign} & Risk of "perverse instantiation" where agents exploit literal, rigid goals with unintended consequences. & Risk of goal drift, prompt hacking, and value misgeneralization where agents pursue correlated but incorrect proxies. & \textbf{Symbolic:} Comprehensive failure mode testing. \textbf{Neural:} Red teaming, constitutional AI, and harmlessness training. \textbf{Universal:} Sandboxed testing environments. \\
\midrule
\textbf{Autonomy vs. Control} \cite{feng2025levels, tallam2025alignment} & Human oversight is typically designed as explicit veto points or permission gates within a deterministic loop. & Human oversight is fuzzy, often implemented as "human-in-the-loop" feedback, which can be ignored or gamed by the agent. & \textbf{Define "meaningful human control" by paradigm.} Symbolic: Clear interrupt signals. Neural: Confidence thresholding for automatic deferral and nuanced steering mechanisms. \\
\midrule
\textbf{Security \& Resilience} \cite{Narajala2025Securing, Khan2024Security} & Vulnerabilities include logic bombs, sensor spoofing, and exploiting algorithmic flaws. & Vulnerabilities include prompt injection, training data poisoning, and adversarial attacks on embeddings. & \textbf{Paradigm-specific defense:} Symbolic: Formal verification, intrusion detection. Neural: Advanced prompt hardening, detection of out-of-distribution inputs, data provenance. \\
\bottomrule
\end{tabular}
}
\label{tab:ethical_challenges}
\end{table}

\vspace{2mm}
\noindent
\textbf{Analysis and Summary}

The bifurcation of ethical challenges detailed in Table \ref{tab:ethical_challenges} leads to several critical and interconnected conclusions. First, it becomes evident that \textbf{effective governance cannot be architecturally agnostic}. Regulation and ethical oversight must be predicated on the underlying paradigm; a requirement for "full explainability," for instance, is feasible for a symbolic system but may be technologically impossible for a pure neural agent, thus necessitating the development of alternative compliance mechanisms.

Furthermore, the rise of \textbf{hybrid systems compounds ethical complexity}. An agent that blends paradigms inherently inherits the governance challenges of both. A neuro-symbolic architecture, for example, requires a framework capable of auditing its deterministic symbolic logic while simultaneously monitoring its neural components for stochastic failures, creating a significantly more demanding oversight burden.

Conversely, \textbf{the attribution gap presents a specific crisis for the neural paradigm}. The fundamental question of "Who is liable?" is most acute here, as its diffuse and stochastic nature directly challenges legal frameworks built on principles of direct causation and intent. This may ultimately require the establishment of new forms of strict liability for developers and operators.

Finally, these distinctions mean that \textbf{effective human-AI collaboration is inherently paradigm-dependent}. Designing appropriate human oversight requires a deep understanding of the agent's core mechanics. The process of overseeing a symbolic agent is analogous to supervising a junior programmer—it involves checking their logical steps. In stark contrast, overseeing a neural agent is more akin to supervising a brilliant but unpredictable intern—it requires carefully steering their context and interpreting their often-opaque outputs.

Addressing the ethical and governance issues of Agentic AI is essential to harness its transformative potential. However, this analysis demonstrates that a nuanced, paradigm-specific approach is not just beneficial but necessary. Blanket policies will inevitably fail. The path forward requires technical standards, legal frameworks, and ethical guidelines that are as sophisticated and differentiated as the technologies they aim to govern. 

This paradigm-specific framing, however, remains incomplete without explicit consideration of policy frameworks that account for the degrees of agency and autonomy in Agentic AI systems, an issue we address next.

\subsection{Toward Agentic AI Policy}
\label{sec:ai_policy}

An overlooked but critical dimension of ethical and governance discourse is the explicit development of \textit{policy frameworks tailored to agentic AI}. Current governance proposals often extend existing AI regulations to cover autonomous systems, but they seldom distinguish between systems that merely generate outputs and those that \textbf{exercise agency in decision-making}. For agentic AI, the challenge lies in \textbf{defining and operationalizing levels of autonomy} and clarifying their governance implications.  

Policy mechanisms must therefore incorporate criteria that distinguish different levels of agency. 
Table \ref{tab:agency_policy} summarizes a proposed taxonomy of agency in Agentic AI, outlining the 
characteristics of assistive, shared, and delegated forms of agency alongside their governance implications.

\begin{table}[htbp]
\centering
\caption{Levels of Agency in Agentic AI and Corresponding Policy Needs}
\resizebox{\textwidth}{!}{
\begin{tabular}{p{3cm}p{6cm}p{6cm}}
\toprule
\textbf{Agency Level} & \textbf{Characteristics} & \textbf{Governance and Policy Requirements} \\
\midrule
\textbf{Assistive} & AI provides recommendations or analysis, with all final decisions made by humans. & Ensure transparency and explainability. Policies should mandate auditability of outputs but allow flexible use with human oversight. \\
\midrule
\textbf{Shared} & AI participates in decision-making, influencing outcomes jointly with human actors. & Require clear role allocation, decision-logging, and mechanisms for tracing contributions of human vs. AI actors. Liability is shared and must be explicitly codified. \\
\midrule
\textbf{Delegated} & AI agents operate with high autonomy, executing decisions or actions within defined domains. & Strong accountability mechanisms, predefined bounds of autonomy, and strict liability regimes for developers/operators. Requires robust monitoring and override capabilities. \\
\bottomrule
\end{tabular}
}
\label{tab:agency_policy}
\end{table}

Accordingly, governance must move beyond paradigm-specific risk analysis toward a \textbf{taxonomy of agency}, where ethical principles and legal accountability mechanisms scale with the degree of autonomy. This aligns with calls for ``meaningful human control'' \cite{feng2025levels}, but extends them into concrete policy design that recognizes the unique governance needs of agentic AI.


\section{Research Gaps: A Paradigm-Specific Roadmap}
\label{sec:gaps}

The development of Agentic AI is constrained by significant, unresolved challenges. However, a critical oversight in identifying these gaps is treating them as uniform across architectures. The research imperatives for symbolic systems diverge profoundly from those for neural systems, with a particularly pressing need for work on hybrid architectures that can leverage the strengths of both. As outlined in Table \ref{tab:research_gaps}, these thematic areas require a paradigm-aware research strategy to ensure future systems are robust, adaptable, and aligned.

\begin{table}[htbp]
\centering
\caption{Paradigm-Specific Research Gaps and Imperatives in Agentic AI}
\resizebox{\textwidth}{!}{
\begin{tabular}{p{3.5cm} p{4.5cm} p{4.5cm} p{4cm}}
\toprule
\textbf{Gap Area} & \textbf{Symbolic Paradigm Challenges} & \textbf{Neural Paradigm Challenges} & \textbf{Research Imperatives} \\
\midrule
\textbf{Evaluation \& Benchmarks} \cite{moshkovich2025beyond, zhuge2024agent} & Lack of standardized metrics for scalability and robustness of logical reasoning in complex, open-world environments. & Current benchmarks (e.g., \textit{AgentBench} \cite{liu2023agentbench}, \textit{GAIA} \cite{Mialon2023GAIA}) fail to adequately test for subtle misalignments, prompt robustness, and the true cost of context management. & Develop paradigm-specific benchmarks. Symbolic: Test logical soundness and failure predictability. Neural: Test for hallucination under pressure, prompt injection resilience, and multi-session consistency. \\
\midrule
\textbf{Reasoning \& Adaptability} \cite{wu2025agentic, zhuang2025self} & Systems are brittle; they fail catastrophically when faced with novel scenarios or exceptions not covered by their rules. & Agents struggle with true, abstract reasoning and value-laden judgment. Their "reasoning" is often just sophisticated pattern matching that can break down. & \textbf{Hybrid Research:} Investigate neuro-symbolic architectures where neural components handle pattern recognition and symbolic modules enforce rigorous reasoning and constraint checking. \\
\midrule
\textbf{Long-term Autonomy \& Memory} \cite{zheng2025lifelong} & Can maintain a persistent, symbolic state but struggle to learn and update their world model from experience in a scalable way. & Context window limitations create agents with severe amnesia across sessions. Statelessness prevents cumulative learning and building long-term relationships. & \textbf{Symbolic:} Research on efficient belief revision. \textbf{Neural:} Develop architectures for external, structured memory that agents can reliably read from and write to. \\
\midrule
\textbf{AI Infrastructure Dependence}  \cite{radanliev2025rise} & Performance is often constrained by the scalability of theorem provers and logic engines, which are sensitive to hardware architecture. Less dependent on massive cloud clusters but requires specialized, reliable compute. & Extreme dependence on vast, expensive cloud compute for training and inference. Creates environmental costs, centralizes power, and creates vulnerabilities to supply chain and geopolitical disruptions. & Develop energy-efficient and decentralized computing paradigms. Research model distillation, sparse architectures, and hybrid cloud-edge deployment to reduce reliance on monolithic infrastructure. \\
\midrule
\textbf{Human-AI Interaction \& Interface Design} \cite{schmidt2023interacting} & Interfaces are typically explicit (e.g., config files, rule editors). The goal is to augment human intelligence with transparent, predictable tools. The distinction between user and agent is clear. & The goal is often a collaborative, conversational partner. Risk of creating opaque "oracles" that users over-trust. Challenges in designing intuitive interfaces for steering, interrupting, and interpreting the stochastic outputs of neural agents. & Establish principles for \textbf{paradigm-aware HCI}. Symbolic: Develop advanced visualization for logic and state. Neural: Research intuitive methods for context steering, confidence communication, and collaborative task management. \\
\midrule
\textbf{Trust \& Transparency} \cite{lakkamraju2025agentic, borghoff2025human} & "How" decisions are made is transparent (the logic trace), but "why" a specific rule exists can be opaque. & Both "how" and "why" are opaque. Explanations are post-hoc and often unreliable. This is the primary barrier to high-stakes deployment. & \textbf{Symbolic:} Research on making goal structures and utility functions explicable. \textbf{Neural:} Fundamental research on mechanistic interpretability and generating faithful, real-time explanations. \\
\midrule
\textbf{Safety \& Alignment} \cite{hellrigel2025misalignment, zhang2025agentalign} & Risk of "perverse instantiation" – perfectly executing a flawed or oversimplified goal specification with catastrophic results. & Vulnerability to prompt injection, goal drift, and value misgeneralization. Aligning a stochastic model to complex human values is an unsolved problem. & \textbf{Paradigm-specific strategies:} Symbolic: Formal verification of goals and constraints. Neural: Advanced red teaming, adversarial training, and "constitutional" oversight mechanisms. \\
\midrule
\textbf{Interoperability \& Integration} \cite{jeong2025study, li2025glue} & Difficult to integrate with the messy, unstructured data of the real world and modern software ecosystems. & Excel at using tools via APIs but struggle with true, semantic understanding of what a tool does, leading to misuse. & Develop standards and middleware for \textbf{paradigm bridging}. Create APIs that allow neural agents to query symbolic reasoners for validation and symbolic systems to leverage neural networks for perception. \\
\midrule
\textbf{Governance \& Accountability} \cite{tennant2023hybrid, rossi2019building} & Liability is more straightforward (flawed logic can be traced) but frameworks for auditing complex rule sets are needed. & A profound "attribution gap" exists. Legal frameworks are unprepared for harm caused by emergent, stochastic behavior. & \textbf{Urgently develop paradigm-specific regulatory models.} Symbolic: Audit trails for decision logic. Neural: Mandatory context logging, output watermarking, and potentially new forms of developer liability. \\
\bottomrule
\end{tabular}
}
\label{tab:research_gaps}
\end{table}

\vspace{2mm}
\noindent
\paragraph{\textbf{Commentary on Key Themes}}

The bifurcation of research gaps identified in Table \ref{tab:research_gaps} reveals that the most critical overarching challenge is the current lack of \textbf{Paradigm-Aware Research Methodologies}. The tools, benchmarks, and success criteria developed for one paradigm frequently prove irrelevant or misapplied to the other, creating fundamental barriers to coherent progress.

This analysis suggests several imperative directions for future work. First, the most promising research path forward appears to lie not in pursuing either paradigm in isolation, but in their \textbf{intentional integration}. The "Reasoning \& Adaptability" gap, for instance, represents a prime candidate for neuro-symbolic solutions, wherein a neural network's robust pattern recognition capabilities are systematically guided and constrained by a symbolic reasoner's logical framework.

Furthermore, the community must move \textbf{beyond isolated benchmarks} that fail to account for paradigmatic differences. There is a critical need to develop separate, rigorous evaluation suites that stress-test the unique failure modes of each architecture—such as logic bombs and edge-case reasoning for symbolic systems, and prompt injection resilience and output stability for neural systems.

Perhaps most urgently, this bifurcation demonstrates that \textbf{effective governance cannot follow a one-size-fits-all approach}. Policymakers and ethicists must collaborate with engineers to develop distinct, tailored frameworks for auditing and regulating these fundamentally different technologies. Applying the stringent verifiability standards of symbolic systems to neural architectures would inadvertently stifle innovation, while applying the more flexible standards designed for neural systems to symbolic environments would overlook critical risks associated with logical integrity and deterministic failure.

\vspace{2mm}
\noindent
\textbf{Conclusion}

Addressing these gaps requires a conscious departure from generic AI research. Progress hinges on a dual-track strategy that deepens our understanding of each paradigm's unique challenges while simultaneously pioneering architectures and standards for their integration. This paradigm-specific roadmap is essential to move from powerful but flawed prototypes to reliable and trustworthy agentic systems. The future of Agentic AI is not a choice between symbolism and connectionism, but a strategic synthesis of both.



\section{Future Directions: The Path to Hybrid Intelligence}
\label{sec:future}

Agentic AI systems are rapidly evolving beyond static task automation into dynamic, collaborative, and adaptive entities \cite{ma2025agentic}. Their future development will hinge on interdisciplinary advances, technological convergence, and—critically—a paradigm-aware approach to design that seeks to integrate the strengths of both symbolic and neural lineages into robust hybrid architectures.

A summary of these paradigm-aware trajectories is presented in Table \ref{tab:future_directions}, which outlines the specific research and integration priorities for each paradigm's evolution, moving beyond a generic technology forecast.

\begin{table}[htbp]
\centering
\caption{Paradigm-Aware Strategic Trajectories for Agentic AI}
\resizebox{\textwidth}{!}{
\begin{tabular}{p{3.5cm}p{4.5cm}p{5cm}} 
\toprule
\textbf{Strategic Direction} & \textbf{Symbolic Paradigm Evolution} & \textbf{Neural Paradigm Evolution} \\
\midrule
\textbf{Multi-Agent Ecosystems} & Defining verifiable communication protocols and interaction contracts for hybrid agent teams. & Specializing in emergent, role-based collaboration and negotiation \cite{huang2025agent, bachrach2020negotiating} (e.g., CrewAI, AutoGen, LangGraph). \\
\midrule
\textbf{Technological Convergence} & Providing the reliable, verifiable logic layer for cyber-physical systems and smart infrastructure. & Acting as the adaptive interface for integrating with IoT, robotics, blockchain, and quantum computing \cite{sultanow2025quantum, radanliev2025rise}. \\
\midrule
\textbf{Self-Evolving Architectures} & Research into automated theorem proving and logical rule discovery for system self-improvement. & Advancing meta-learning and feedback-driven optimization \cite{ma2025agentic} for architecture tuning and deployment-aware adaptation. \\
\midrule
\textbf{Human-AI Collaboration} & Enabling interfaces where humans can directly inspect, debug, and modify an agent's logical rule set and goals. & Creating intuitive interfaces for shared intent and cognitive/emotional responsiveness \cite{schmidt2023interacting} via natural language. \\
\midrule
\textbf{Governance-First Design} & Formal verification of goal structures and safety constraints for embeddable governance modules. & Developing techniques for embedded ethics, policy enforcement, and global accountability \cite{gabison2025inherent} within stochastic systems (e.g., IBM Governance Stack). \\
\midrule
\textbf{Scientific Discovery} & Encoding scientific laws and methodological rigor for agent-led hypothesis generation. & Driving agent-led inquiry and results analysis \cite{koutra2025towards, lu2024ai} in platforms like Sakana AI Scientist \cite{lu2024ai} and Microsoft Discovery. \\
\midrule
\textbf{Research Priorities} & Establishing benchmarks for logical soundness, verifiability, and interoperability standards. & Establishing metrics for moral alignment, cognitive modeling, and alignment \cite{cervantes2020toward} (e.g., AgentBench \cite{liu2023agentbench}). \\
\bottomrule
\end{tabular}
}
\label{tab:future_directions}
\end{table}

\textbf{Analysis of Strategic Trajectories}

The bifurcated future outlined in Table \ref{tab:future_directions} leads to one overriding conclusion: the paramount direction is \textbf{Architectural Integration}. The goal is to forge a new class of hybrid systems that leverage the reliability of symbolic reasoning and the adaptability of neural generation.

\begin{itemize}
    \item \textbf{Neuro-Symbolic Integration as the Keystone:} The most profound progress will come from research that successfully couples neural networks for perception and pattern recognition with symbolic engines for reasoning and constraint checking. This is the most promising path to overcoming the brittleness of pure symbolism and the opacity of pure neural approaches.
    \item \textbf{Paradigm-Specialized Roles in Ecosystems:} Future multi-agent ecosystems \cite{huang2025agent, bachrach2020negotiating} will not be homogenous. They will consist of specialized agents—some highly neural for creative tasks, some highly symbolic for regulatory compliance—that communicate through standardized protocols. The orchestration of such hybrid swarms is a critical research frontier.
    \item \textbf{A Dual-Track Approach to Governance:} The development of safety and governance mechanisms \cite{gabison2025inherent} must continue on two tracks: advancing formal methods for symbolic verifiability \textit{and} developing new statistical, training-based methods for neural alignment. The ultimate governance framework for a hybrid agent will need to seamlessly combine both.
    \item \textbf{Convergence as Amplification:} The integration with other technologies \cite{sultanow2025quantum, radanliev2025rise} will amplify the capabilities of both paradigms. Neural agents will manage real-time sensor data from IoT, while symbolic modules will ensure the decisions made from that data are safe and compliant.
\end{itemize}

\vspace{2mm}
\noindent
\textbf{Conclusion}

The future of Agentic AI is a synthesis. Its trajectory will be shaped not only by technical breakthroughs but by thoughtful, paradigm-aware integration of ethics, interdisciplinary methods, and infrastructure-aware governance \cite{gabison2025inherent}. The next conceptual turning point will be defined by our ability to engineer \textbf{hybrid intelligence}—systems that are both \textit{adaptable} and \textit{reliable}, both \textit{creative} and \textit{sound}. The question is no longer whether agents will become intelligent partners, but whether we can architect a future of hybrid intelligence that is both powerful and trustworthy.


\section{Conclusion}
\label{sec:conclusion}
Agentic AI represents a fundamental paradigm shift in the design of intelligent systems, but its rapid evolution has led to a fragmented and often anachronistic understanding of the field. This review has addressed this confusion by introducing and validating a novel conceptual framework: the existence of two distinct lineages of Agentic AI---the \textbf{Symbolic/Classical} and the \textbf{Neural/Generative}---each with fundamentally different operational mechanics, strengths, and limitations.

Our analysis demonstrates that the common practice of \textit{conceptual retrofitting}---describing modern LLM-orchestrated agents with the language of symbolic systems (e.g., PPAR loops, BDI)---obscures their true nature and impedes progress. Through a systematic, paradigm-aware review of the literature, we have established three central tenets. First, \textbf{the architectural divide is both real and meaningful}; symbolic systems excel in environments requiring safety, verifiability, and explicit logic (e.g., healthcare, robotics control), while neural systems thrive in domains requiring adaptability, pattern recognition, and operation on unstructured data (e.g., finance, creative research) (Sections~\ref{sec:applications},~\ref{sec:taxonomy}).

Furthermore, this divide dictates that \textbf{governance must be paradigm-specific}. The ethical challenges and requisite mitigation strategies differ profoundly between paradigms, meaning accountability for a symbolic system involves auditing its logic, whereas for a neural system, it necessitates auditing its training data and prompts. This renders a one-size-fits-all approach to AI ethics fundamentally insufficient (Section~\ref{sec:ethics}).

Critically, our findings indicate that \textbf{the most productive path forward is hybrid, not isolated}. The most pressing research gaps and promising future directions lie not in the isolated improvement of either paradigm, but in their strategic integration into neuro-symbolic architectures that leverage the complementary strengths of symbolic reliability and neural adaptability (Sections~\ref{sec:gaps},~\ref{sec:future}).

This dual-paradigm framework provides the essential analytical lens to move the field beyond a simple catalog of technologies toward a coherent theory of architectural design in Agentic AI. It offers researchers, engineers, and policymakers a precise vocabulary and a functional taxonomy to classify systems, evaluate their capabilities and risks appropriately, and make informed design choices.

Ultimately, the development of Agentic AI is not merely a technical challenge—it is a sociotechnical one. Its success will depend on whether we can architect systems that are not only powerful but also trustworthy. This requires a conscious and deliberate effort to build hybrid intelligence—systems that are both adaptable and reliable, both creative and sound. By recognizing and embracing the distinct nature of these two architectural lineages, we can steer this transformative technology toward a future where agentic systems truly serve as trusted collaborators in scientific discovery (understanding), in providing fair and accessible services (equity), and in forming the robust, verifiable backbone of critical infrastructure (resilience).




\bibliographystyle{unsrt}

\end{document}